\documentclass[conference]{IEEEtran}
\usepackage{graphics}
\usepackage{graphicx}
\begin{document}
\title{Simultaneous Bayesian inference of motion velocity fields 
and probabilistic models in successive video-frames described by spatio-temporal MRFs}
\author{\IEEEauthorblockN{Yuya Inagaki and Jun-ichi Inoue}
\IEEEauthorblockA{Complex Systems Engineering, Graduate School of Information Science and Technology\\
Hokkaido University, N14-W-9, Kita-ku, Sapporo 060-0814, Japan \\ 
Email: inagaki@chaos1.complex.eng.hokudai.ac.jp, j$\underline{\,\,\,}$inoue@complex.eng.hokudai.ac.jp}}
\maketitle
\begin{abstract}
We numerically investigate a mean-field Bayesian approach 
with the assistance of the Markov chain Monte Carlo method 
to estimate motion velocity fields  
and probabilistic models simultaneously in consecutive digital images 
described by spatio-temporal Markov random fields. 
Preliminary to construction of  our procedure, we find that mean-field variables in the iteration diverge 
due to improper normalization factor of regularization terms appearing 
in the posterior. To avoid this difficulty, we rescale the regularization 
term by introducing a scaling factor and optimizing it by means of minimization 
of the mean-square error. We confirm that the optimal scaling factor 
stabilizes the mean-field iterative process of the motion velocity 
estimation. We next attempt to estimate the optimal values of 
hyper-parameters including the regularization term, which define 
our probabilistic model macroscopically, by using the Boltzmann-machine type 
learning algorithm based on gradient descent of 
marginal likelihood (type-II likelihood) with respect to the hyper-parameters. 
In our framework, one can estimate both the probabilistic model (hyper-parameters) 
and motion velocity fields simultaneously.  
We find that our motion estimation is much better than the result obtained 
by Zhang and Hanouer (1995) in which the hyper-parameters are set to some ad-hoc values 
without any theoretical justification.
\end{abstract}
\IEEEpeerreviewmaketitle
\section{Introduction}
\label{sec:Intro}
Motion estimation in 
consecutive video-frames is 
one of the important techniques in 
image processing or computer vision community. 
The motion estimation is defined as 
estimating the motion velocity fields 
(vectors) of objects appearing in 
successive two (video) frames. 
In the research field of computer vision, 
the so-called 
Markov random fields (MRFs for short) 
have been used to 
solve the various problems concerning 
image processing 
such as image restoration \cite{Geman}, texture analysis and 
segmentation \cite{Netravali,Winkler,Bishop}, super-resolution 
\cite{Park,Kanemura} and so on. 
The MRFs enable us to regularize 
the ill-posed problems in such a lots of 
subjects,  and then, the original problem can be treated as 
combinatorial optimization problems 
under some `soft' or `hard' constraints. 
Actually, Zhang and Hanouer (1995) \cite{Zhang} and 
Wei and Li (1999) \cite{Wei} 
applied the MRFs approach with the assistance of 
the framework of 
Bayesian statistics to 
estimate the motion vector for a 
given two consecutive digital images. 
They also utilized the so-called mean-field approximation 
to carry out the extensive sums 
in the marginal probability of the posterior and showed that 
the steady states of the mean-field equations 
are one of the good candidates for the appropriate motion velocity fields. 
The same kind of the MRFs approach 
was implemented by making use of 
the DSP-based image processing board of 
SIMD (Single Instruction Multiple Data) machine 
by Caplier, Luthon and Dumontier (1998) \cite{Caplier} and 
Luthon, Caplier and Lievin (1999) \cite{Luthon}. They demonstrated 
that the task to estimate the motion velocity is actually carried out within a realistic time.  

In the study by Zhang and Hanouer (1995), 
they set the so-called hyper-parameters which 
specify the probabilistic model 
macroscopically to some ad-hoc values without any reasonable explanation. 
However,  there is no theoretical  (statistical) justification 
for such ad-hoc choices of parameters to estimate the appropriate motion velocity fields. 
Of course, the selection of hyper-parameters 
is dependent on a given set of consecutive video-frames 
and it is important for us to 
determine the hyper-parameters systematically 
under some statistical criteria so as to 
give a fine (if possible, an optimal) 
average-case performance of the motion estimation. 

Taking into account the above requirements from both theoretical and 
practical sides, 
from the view point of Bayesian statistics, we examine a mean-field approach 
with the assistance of the Markov chain Monte Carlo method (the MCMC for 
short) to estimate both motion velocity fields 
and hyper-parameters simultaneously in successive video-frames described by 
spatio-temporal MRFs. We find that mean-field variables in the non-linear 
maps diverge due to improper normalization factor of regularization terms 
appearing in the cost function. In order to overcome this difficulty, 
we rescale the regularization terms by introducing a scaling factor and optimizing it 
by means of minimization of the mean-square error.  We reveal that the optimal scaling factor 
stabilizes the mean-field iterative procedure of the motion 
velocity fields estimation. We next attempt to estimate the optimal values of 
hyper-parameters including the regularization term, which define 
our probabilistic model macroscopically, by using the {\it Boltzmann-machine} type   
learning algorithm based on gradient descent of the marginal likelihood 
with respect to hyper-parameters. 
In our framework, one can estimate both the probabilistic model (hyper-parameters) 
and motion fields simultaneously. We show that our motion estimation 
is much better than the result given by Zhang and Hanouer (1995) 
in which hyper-parameters are set to some ad-hoc values 
without any theoretical explanation.

This paper is organized as follows. 
In the next section \ref{sec:set-up}, 
we explain our general set-up to deal with the motion velocity 
estimation by means of spatio-temporal 
MRFs according to Zhang and Hanouer (1995). 
From the view point of Bayesian 
inference, we construct the posterior probability 
and introduce two kinds of estimations, namely, 
Maximum A Posteriori (MAP for short)  and 
Maximizer of Posterior Marginal (MPM for short) estimations. 
In section \ref{sec:MFA}, 
we utilize the mean-field approximation 
to obtain the MPM estimate and 
derive the non-linear mean-field equations 
with respect to the motion velocity fields. 
As a preliminary,  we demonstrate our mean-field approach 
by setting the hyper-parameters to the values 
chosen by Zhang and Hanouer (1995) and 
show that the mean-fields diverge leading up to 
a quite worse estimation of motion velocity in section \ref{sec:PRE}. 
To avoid this type of difficulty, we shall rescale the regularization 
term by introducing a scaling factor and optimizing it by means of minimization 
of the mean-square error. 
In section \ref{sec:ML}, 
we attempt to estimate the optimal values of 
hyper-parameters including the regularization term, which define 
our probabilistic model macroscopically, by using the Boltzmann-machine type 
learning algorithm based on gradient descent 
of the marginal likelihood with respect to hyper-parameters. 
In our framework, one can estimate both the probabilistic model (hyper-parameters) 
and motion velocity fields simultaneously. 
To proceed to solve the learning 
equations, we utilize two different ways to carry out the sums 
coming up exponential order 
appearing in the learning equations, 
namely, 
hybridization of mean-field approximation and MCMC, 
and simple MCMC. 
We find that average-case performance of our motion 
estimation is 
much better than the result given by Zhang and Hanouer (1995) 
in which the hyper-parameters 
are set to some ad-hoc values.
The last section is summary.
\section{General set-up of motion estimation}
\label{sec:set-up}
In this section, we briefly explain 
our model system. 
\subsection{Spatio-temporal Markov random fields}
Let us define a 
single two-dimensional gray-scale 
image as a `video-frame' 
by $\mbox{\boldmath $x$}^{\tau}= 
\{x_{i}^{\tau}, i\in S\}$. 
$S$ denotes a set of 
pixels in image and index $i$ is related to 
a point in two-dimensional square lattice 
$(x,y)$.  Here we shall assume that 
a motion picture consists of 
successive static images (frames), 
namely, we distinguish each static image 
in the motion picture by time index $\tau$ as 
$\mbox{\boldmath $x$}^{\tau}$. 
When we compare the consecutive two 
static images, 
that is, $\mbox{\boldmath $x$}^{\tau-1}$ and 
$\mbox{\boldmath $x$}^{\tau}$, 
each pixel in 
$\mbox{\boldmath $x$}^{\tau}$ 
might change its location with some `motion velocity'. 
From this assumption in mind, we 
introduce velocity fields  
defined by $\mbox{\boldmath $d$}^{\tau} = 
\{d_{i}^{\tau}, i \in S\}$. 
Namely, 
for each $i$ and for successive two video-frames, 
a constraint 
$x_{i}^{\tau}   =   x_{i-d_{i}^{\tau}}^{\tau-1}$ 
should be satisfied, where `index' $d_{i}^{\tau}$ 
is related to a  single 
point $(v_{x}^{\tau}(i),v_{y}^{\tau}(i))$ in 
the two-dimensional vector field. 
In this paper, we consider that 
each component of the vector takes a 
discrete value and  the range is limited as 
$|v_{x}^{\tau}(i)|, |v_{y}^{\tau}(i)| \leq d_{\rm max}-1 = 5$. 
It might seem that this range is extremely small in 
comparison with the range of 
the grayscales in images (from $0$ to 
$255$) or image size ($\sim 30 \times 30$), 
however, if one attempts to 
construct 
a detection and alarming system for the dangerous state 
from `infinitesimal difference' of patient's breath in ICU (Intensive Care Unit), 
the limitation of the velocity fields to such a small range 
is rather desirable (reasonable). 
\subsubsection{Line fields and segmentation fields}
Obviously, 
it is impossible to 
determine the 
$\mbox{\boldmath $d$}^{\tau} = 
\{d_{i}^{\tau}, i \in S\}$ uniquely
from just only information about 
two video-frames 
$\mbox{\boldmath $x$}^{\tau}$ and 
$\mbox{\boldmath $x$}^{\tau-1}$. 
To compensate this lack information, we introduce 
line fields and segmentation fields. 

The {\it line fields} guarantee 
the continuousness between arbitrary two motion velocity 
fields  for the nearest neighboring pixels and 
we assume that  these two motion velocity fields  
might take similar values. 
Let us define these line fields 
by $\mbox{\boldmath $l$}=\{l(i,j)|l(i,j) \equiv  
(h_{i},v_{i},h_{j},v_{j}) \in S\}$. 
Here $h_{i}$ and 
$v_{i}$ are 
labels to represent 
continuousness between 
velocity fields  in the nearest neighboring (n.n. for short) 
horizontal and vertical pixels. 
In other words, we shall define  
\begin{eqnarray*}
h_{i}^{\tau}  & = &   \left\{ \begin{array}{ll}
0 & \mbox{($\mbox{\boldmath $d$}^{\tau}$s for  
horizontally n.n. pixels are discont.)} \\
1 & \mbox{($\mbox{\boldmath $d$}^{\tau}$s for  
horizontally n.n. pixels are cont.)} \\
\end{array} \right. \\
v_{i}^{\tau} & = &   \left\{ \begin{array}{ll}
0 & \mbox{($\mbox{\boldmath $d$}^{\tau}$s for  
vertically n.n. pixels are discont.)} \\
1 & \mbox{($\mbox{\boldmath $d$}^{\tau}$s for  
vertically n.n. pixels are cont.)} \\
\end{array} \right. 
\end{eqnarray*} 
\mbox{}

On the other hand, 
the {\it segmentation fields} are 
introduced to 
distinguish `predictable areas'  
and `unpredictable areas' in 
the motion velocity fields. 
Here `unpredictable areas' 
means regions hided by some objects before 
they are moving to somewhere else. 
Thus, we naturally define 
the segmentation fields by $\mbox{\boldmath $s$}=\{s_{i}|s_{i} =0,1 \}$ with 
\begin{eqnarray*}
s_{i}^{\tau} & = & \left\{ \begin{array}{ll}
 0 & \mbox{(pixel $i$ is predictable)}\\
 1 & \mbox{(pixel $i$ is unpredictable)}\\
\end{array} \right.
\end{eqnarray*}
\subsection{Bayes rule and posterior probability}
In the previous subsections, we defined 
the motion picture as a series of successive static images 
by spatio-temporal Markov random fields. 
To determine the motion velocity fields uniquely,  
we also introduced the line and segmentation fields. 
Then, our problem is clearly defined as follows. 

Now, our problem is to infer the 
velocity vector field 
$\mbox{\boldmath $d$}^{\tau}$, 
line field $\mbox{\boldmath $l$}^{\tau}$ and 
segmentation field 
$\mbox{\boldmath $s$}^{\tau}$ 
under the condition that 
two consecutive video-images 
$\mbox{\boldmath $x$}^{\tau}$ and 
$\mbox{\boldmath $x$}^{\tau-1}$ are observed.
For the above problem, we easily use the Bayes 
rule to obtain the posterior probability, which is 
a probability of  $\mbox{\boldmath $\Sigma$}^{\tau} \equiv 
\{
\mbox{\boldmath $d$}^{\tau}, 
\mbox{\boldmath $s$}^{\tau},
\mbox{\boldmath $l$}^{\tau}
\}$ 
provided that $\mbox{\boldmath $x$}^{\tau}$ and 
$\mbox{\boldmath $x$}^{\tau-1}$ are given as 
\begin{eqnarray}
P(\mbox{\boldmath $\Sigma$}^{\tau}|
\mbox{\boldmath $x$}^{\tau},
\mbox{\boldmath $x$}^{\tau-1}) &  = &  
\frac{P(\mbox{\boldmath $x$}^{\tau}|
\mbox{\boldmath $\Sigma$}^{\tau},
\mbox{\boldmath $x$}^{\tau-1})
P(\mbox{\boldmath $\Sigma$}^{\tau}|
\mbox{\boldmath $x$}^{\tau-1})}
{\sum_{\mbox{\boldmath $\Sigma$}^{\tau}}
P(\mbox{\boldmath $x$}^{\tau}|\mbox{\boldmath $\Sigma$}^{\tau},
\mbox{\boldmath $x$}^{\tau-1})
P(\mbox{\boldmath $\Sigma$}^{\tau}|\mbox{\boldmath $x$}^{\tau-1})} \nonumber \\
\mbox{} & = &  
\frac{P(\mbox{\boldmath $x$}^{\tau}|
\mbox{\boldmath $\Sigma$}^{\tau},
\mbox{\boldmath $x$}^{\tau-1})
P(\mbox{\boldmath $\Sigma$}^{\tau}|
\mbox{\boldmath $x$}^{\tau-1})}
{P(\mbox{\boldmath $x$}^{\tau}|
\mbox{\boldmath $x$}^{\tau-1})}
\label{eq:beyes}
\end{eqnarray}
where we defined the sums appearing in the above formula by 
$\sum_{\mbox{\boldmath $\Sigma$}^{\tau}}(\cdots) \equiv  
\sum_{\mbox{\boldmath $d$}^{\tau}}(\cdots)
\sum_{\mbox{\boldmath $s$}^{\tau}}(\cdots)
\sum_{\mbox{\boldmath $l$}^{\tau}}(\cdots)$ with 
\begin{eqnarray}
\sum_{\mbox{\boldmath $d$}^{\tau}} (\cdots) &  \equiv & 
\prod_{i=1}^{N}
\sum_{d_{i}=0}^{d_{\rm max}-1}(\cdots) \\
\sum_{\mbox{\boldmath $s$}^{\tau}} (\cdots) &  \equiv & 
\prod_{i=1}^{N}
\sum_{s_{i}=0,1}(\cdots) \\
\sum_{\mbox{\boldmath $l$}^{\tau}} (\cdots) & \equiv &  
\prod_{i=1}^{N} 
\sum_{h_{i}=0,1}
\sum_{v_{i}=0,1}(\cdots). 
\end{eqnarray}
For the above posterior, we have the so-called 
{\it Maximum A Posteriori (MAP)} estimate by 
\begin{eqnarray}
\mbox{\boldmath $\Sigma$}_{MAP}^{\tau} & = & 
\arg\max_{\mbox{\boldmath $\Sigma$}^{\tau}}
\log 
P(\mbox{\boldmath $\Sigma$}^{\tau}|
\mbox{\boldmath $x$}^{\tau},
\mbox{\boldmath $x$}^{\tau-1})
\label{eq:def_MAP}
\end{eqnarray}
whereas,  what we call  
{\it Maximizer of Posterior Marginal (MPM)} estimate is 
given by 
\begin{eqnarray}
\Sigma_{i, MPM}^{\tau} & = & 
\arg\max_{\Sigma_{i}^{\tau}}
P(\Sigma_{i}^{\tau}|
\mbox{\boldmath $x$}^{\tau},
\mbox{\boldmath $x$}^{\tau-1}) = 
Q(\langle \Sigma_{i}^{\tau} \rangle)
\label{eq:def_MPM}
\end{eqnarray}
where we defined the marginal probability by 
\begin{eqnarray}
 P(\Sigma_{i}^{\tau}|
\mbox{\boldmath $x$}^{\tau},
\mbox{\boldmath $x$}^{\tau-1})  & \equiv & 
\sum_{\mbox{\boldmath $\Sigma$}^{\tau} \neq 
\Sigma_{i}^{\tau}}
P(\mbox{\boldmath $\Sigma$}^{\tau}|
\mbox{\boldmath $x$}^{\tau},
\mbox{\boldmath $x$}^{\tau-1}).
\end{eqnarray}
The average $\langle \cdots \rangle$ appearing in (\ref{eq:def_MPM}) 
is defined as 
$\langle \cdots 
\rangle \equiv 
\sum_{\mbox{\boldmath $\Sigma$}^{\tau}}
(\cdots)
P(\mbox{\boldmath $\Sigma$}^{\tau}|
\mbox{\boldmath $x$}^{\tau},
\mbox{\boldmath $x$}^{\tau-1})$
and $Q(\cdots)$ denotes a function 
to convert the expectation 
$\sum_{\mbox{\boldmath $\Sigma$}^{\tau}}
\mbox{\boldmath $\Sigma$}^{\tau}
P(\mbox{\boldmath $\Sigma$}^{\tau}|
\mbox{\boldmath $x$}^{\tau},
\mbox{\boldmath $x$}^{\tau-1})$ having a 
real number into the nearest discrete value. 
\subsubsection{Likelihood function}
The likelihood function appearing in the posterior 
$P(\mbox{\boldmath $x$}^{\tau}|
\mbox{\boldmath $\Sigma$},
\mbox{\boldmath $x$}^{\tau-1})$ 
can be regarded as a probabilistic model to 
generate the next  frame $\mbox{\boldmath $x$}^{\tau}$ 
provided that 
the unknown fields 
$\mbox{\boldmath $\Sigma$}$ and 
the frame in the previous time 
$\mbox{\boldmath $x$}^{\tau}$ are given.  
From now on, we omit the $\tau$-dependence of 
the fields because we consider 
the motion velocity fields for a given set of 
just only two consecutive video-frames. 
Then, we assume $P(\mbox{\boldmath $x$}^{\tau}|
\mbox{\boldmath $\Sigma$},
\mbox{\boldmath $x$}^{\tau-1}) \propto  
{\exp}
\left[
-E^{(1)}(\mbox{\boldmath $x$}^{\tau}|
\mbox{\boldmath $\Sigma$},
\mbox{\boldmath $x$}^{\tau-1})
\right]$
where the cost function 
$E^{(1)}(\mbox{\boldmath $x$}^{\tau}|
\mbox{\boldmath $\Sigma$},
\mbox{\boldmath $x$}^{\tau-1})$
is given by 
\begin{eqnarray}
E^{(1)}(\mbox{\boldmath $x$}^{\tau}|
\mbox{\boldmath $\Sigma$},
\mbox{\boldmath $x$}^{\tau-1})  & = &  
\frac{1}{2\sigma^{2}}\sum_{i}
(1-s_{i})(x_{i}^{\tau}
-x_{i-d_{i}}^{\tau-1})^{2} \nonumber \\
\mbox{} & + &  
\alpha_{l}
\sum_{i,j\in \mbox{\boldmath $N$}(i)}
\frac{l(i,j)}{(x_{i}^{\tau}-x_{j}^{\tau})^{2}} 
\end{eqnarray}
where $\mbox{\boldmath $N$}(i)$ 
means a set of 
nearest neighboring 
pixels around pixel $i$. 
The number of 
these pixels is 
$|\mbox{\boldmath $N$}(i)|=4$ (square lattice).  
The parameters $\sigma$ and 
$\alpha_{l}$ are the so-called {\it hyper-parameters} 
which determine the 
probabilistic model macroscopically. 
\subsubsection{Prior probability}
The prior probability 
$P(\mbox{\boldmath $\Sigma$}|\mbox{\boldmath $x$}^{\tau})$ 
is a generating model of 
the fields 
$\mbox{\boldmath $\Sigma$}^{\tau}$ for 
a given frame $\mbox{\boldmath $x$}^{\tau}$ 
and it is given by $P(\mbox{\boldmath $\Sigma$}|
\mbox{\boldmath $x$}^{\tau}) \propto  
{\exp}
\left[
-E^{(2)}
(\mbox{\boldmath $\Sigma$}| 
\mbox{\boldmath $x$}^{\tau})
\right]$ 
with 
\begin{eqnarray}
E^{(2)}(\mbox{\boldmath $\Sigma$}| 
\mbox{\boldmath $x$}^{\tau})  &&  \hspace{-0.8cm} =   
\lambda_{d}\sum_{i,j\in \mbox{\boldmath $N$}(i)}
(1-2\,{\rm e}^{-\beta_{d} \parallel d_{i}-d_{j} \parallel^{2}})
(1-l(i,j)) \nonumber \\
\mbox{} & + & 
\lambda_{s}\sum_{i,j\in \mbox{\boldmath $N$}(i)}(1-l(i,j))
(1-2\delta(s_{i}-s_{j})) \nonumber \\
\mbox{} & + &  T_{s}
\sum_{i}s_{i}
\label{eq:energy}
\end{eqnarray}
where 
we defined the norm 
$\parallel \cdots \parallel$ by 
\begin{eqnarray*}
\parallel d_{i} -d_{j} \parallel & \equiv & 
\left\{
(v_{x}(i)-v_{x}(j))^{2}+
(v_{y}(i)-v_{y}(j))^{2}
\right\}^{1/2}
\end{eqnarray*}
and 
$\lambda_{d}, \lambda_{s}, \lambda_{l}, \beta_{d}$ and $T_{s}$ 
are also hyper-parameters which define 
the above probabilistic model macroscopically. 
\subsubsection{Posterior}
Then, the posterior $P(\mbox{\boldmath $\Sigma$} | 
\mbox{\boldmath $x$}^{\tau},
\mbox{\boldmath $x$}^{\tau-1})$, 
namely, 
the probability 
of the desired fields 
for a given set of 
two successive video-frames 
$\mbox{\boldmath $x$}^{\tau},
\mbox{\boldmath $x$}^{\tau-1}$ 
is constructed by the product of 
likelihood $P(\mbox{\boldmath $x$}^{\tau}|
\mbox{\boldmath $\Sigma$},
\mbox{\boldmath $x$}^{\tau-1})$ 
and prior 
$P(\mbox{\boldmath $\Sigma$}|
\mbox{\boldmath $x$}^{\tau-1})$, 
that is  
$P(\mbox{\boldmath $\Sigma$}| 
\mbox{\boldmath $x$}^{\tau},
\mbox{\boldmath $x$}^{\tau-1}) \propto 
P(\mbox{\boldmath $x$}^{\tau}|
\mbox{\boldmath $\Sigma$},
\mbox{\boldmath $x$}^{\tau-1})
P(\mbox{\boldmath $\Sigma$}|
\mbox{\boldmath $x$}^{\tau-1})$. 

By means of the cost function, we have 
\begin{eqnarray}
P(\mbox{\boldmath $\Sigma$}| 
\mbox{\boldmath $x$}^{\tau},
\mbox{\boldmath $x$}^{\tau-1}) & \propto & 
{\exp}
\left[
-E^{(1)}(\mbox{\boldmath $x$}^{\tau}|
\mbox{\boldmath $\Sigma$},
\mbox{\boldmath $x$}^{\tau-1}) 
-
E^{(2)}(\mbox{\boldmath $\Sigma$}| 
\mbox{\boldmath $x$}^{\tau}) 
\right] \nonumber \\
\mbox{} & \equiv  & 
{\exp}
\left[
-E
(\mbox{\boldmath $\Sigma$}| 
\mbox{\boldmath $x$}^{\tau},
\mbox{\boldmath $x$}^{\tau-1})
\right]. 
\end{eqnarray}
The total cost of the system, which is now defined by 
$-\log P(\mbox{\boldmath $\Sigma$} | 
\mbox{\boldmath $x$}^{\tau},
\mbox{\boldmath $x$}^{\tau-1})$, is written as 
\begin{eqnarray}
&& 
E(\mbox{\boldmath $\Sigma$} | 
\mbox{\boldmath $x$}^{\tau},
\mbox{\boldmath $x$}^{\tau-1})  \equiv   
\frac{1}{2\sigma^{2}}\sum_{i}
(1-s_{i})(x_{i}^{\tau}
-x_{i-d_{i}}^{\tau-1})^{2} \nonumber \\
\mbox{} & + &  
\lambda_{d}\sum_{i,j\in \mbox{\boldmath $N$}(i)}
(1-2\,{\rm e}^{-\beta_{d} \parallel d_{i}-d_{j} \parallel^{2}})
(1-l(i,j)) \nonumber \\
\mbox{} & + & 
\lambda_{s}\sum_{i,j\in \mbox{\boldmath $N$}(i)}(1-l(i,j))
(1-2\delta(s_{i}-s_{j})) \nonumber \\
\mbox{} & + &  
\alpha_{l}
\sum_{i,j\in \mbox{\boldmath $N$}(i)}
\frac{l(i,j)}{(x_{i}^{\tau}-x_{j}^{\tau})^{2}}  + T_{s}\sum_{i}s_{i}
\label{eq:energy}
\end{eqnarray}
where 
the first term appearing 
in the right hand side of the above cost function 
is introduced 
to prevent pixel $x_{i}^{\tau-1}$ at the location $i$  
from moving to the position 
$i-d_{i}^{\tau}$ where is quite far from $i$. 
The second term confirms the continuousness 
between velocity vectors for the nearest neighboring pixels and we easily find  
that the term is identical to 
the Hamiltonian (energy function) for the so-called 
dynamically diluted {\it ferromagnetic Q-Ising model} 
in the literature of statistical physics, that is 
to say, we have 
\begin{eqnarray}
&& \hspace{-2cm}
\lambda_{d}
\sum_{i,j \in \mbox{\boldmath $N$}(i)}
(1-l(i,j))
(1-2\,{\rm e}^{-\beta_{d} \parallel d_{i}-d_{j} \parallel^{2}}) \nonumber \\
\mbox{} & \simeq & 
2\lambda_{d}\beta_{d}
\sum_{i,j \in \mbox{\boldmath $N$}(i)}
(1-l(i,j)) \parallel d_{i}-d_{j} \parallel^{2} \nonumber \\
\mbox{} & + &  
\mbox{$\mbox{\boldmath $d$}$-independent const.}
\end{eqnarray}
in the limit of $\beta_{d} \to 0$. 
The third term in (\ref{eq:energy}) 
denotes a correlation 
between the line and the segmentation 
fields. 
The forth term 
represents a correlation 
between the line fields and 
the distance of pixels located in the nearest neighboring positions. 
The last term controls the number of 
non-zero segmentation fields  and 
this term can be regarded as the so-called {\it chemical potential} 
in the literature of statistical physics. 
\section{Mean-field equations on pixel}
\label{sec:MFA}
In the previous section, we constructed  
the posterior by making use of the Bayes rule. 
Therefore, we can use both 
MAP and MPM 
estimations by means of (\ref{eq:def_MAP}) 
and (\ref{eq:def_MPM}), respectively. 
Here we should notice that 
the MAP estimate is recovered 
by means of 
\[
\Sigma_{i,MAP} = 
\lim_{\beta \to \infty}
Q(\langle \Sigma_{i} \rangle_{\beta}),\,
\langle \cdots \rangle_{\beta} \equiv  
\sum_{\mbox{\boldmath $\Sigma$}}
(\cdots)
P_{\beta}(\mbox{\boldmath $\Sigma$}|
\mbox{\boldmath $x$}^{\tau},
\mbox{\boldmath $x$}^{\tau-1}) 
\]
with $P_{\beta}(\mbox{\boldmath $\Sigma$}|
\mbox{\boldmath $x$}^{\tau},
\mbox{\boldmath $x$}^{\tau-1}) \
\propto 
{\exp}
\left[
-\beta E
(\mbox{\boldmath $\Sigma$}| 
\mbox{\boldmath $x$}^{\tau},
\mbox{\boldmath $x$}^{\tau-1})
\right]$. 
From the above definitions, 
the MPM estimate is obtained by 
$\Sigma_{i,MPM}=
Q(\langle \Sigma_{i} \rangle_{1})$. 
Therefore, 
our problem now seems to be 
completely solved. However, 
the number of sums appearing in the expectation 
$\langle \cdots \rangle_{\beta}$ 
\begin{eqnarray}
\sum_{\mbox{\boldmath $\Sigma$}}
(\cdots)  & = &   
\sum_{s_{1}=0,1}
\cdots
\sum_{s_{N}=0,1}
\sum_{d_{1}=0}^{d_{\rm max}-1}
\cdots
\sum_{d_{N}=0}^{d_{\rm max}-1}  \nonumber \\
\mbox{} & \times & 
\sum_{l_{1}=0,1}
\cdots
\sum_{l_{N}=0,1}
(\cdots)
\end{eqnarray}
comes up to exponential order as ${\rm e}^{N\log 4d_{\rm max}}$. 
Obviously, it is impossible for us to 
carry out the sums even for the system size is  
$N=30 \times 30=900$ within a realistic time. 

Then, we use the mean-field approximation 
to overcome this type of computational difficulties. 
Namely, we rewrite the 
cost function 
by replacing the motion velocity fields with 
the corresponding expectations except for 
a single component of the fields.
For instance, for say $s_{i}$, we have the mean-field approximated 
cost function as follows.
\begin{eqnarray*}
E & \simeq & E^{0}(s_{i}) 
\equiv  
-\frac{s_{i}}{2\sigma^{2}}
(x_{i}^{\tau}-x_{i-\langle d_{i} \rangle_{\beta}^{\rm mf}}^{\tau-1})^{2} 
+T_{s} s_{i} \nonumber \\
\mbox{} & + &  
\lambda_{s} \sum_{j \in 
\mbox{\boldmath $N$}(i)}
(1-\langle l(i,j) \rangle_{\beta}^{\rm mf})
(1-\delta(s_{i}-\langle s_{j} \rangle_{\beta}^{\rm mf}))
\end{eqnarray*}
By using the same way as $s_{i}$, we have 
for $d_{i}$ as 
\begin{eqnarray*}
E & \simeq & E^{0}(d_{i}) 
\equiv  
\frac{(1-\langle s_{i} \rangle_{\beta}^{\rm mf})}{2\sigma^{2}}
(x_{i}^{\tau}-x_{i-d_{i}}^{\tau-1})^{2} \nonumber \\
\mbox{} & + &   
\lambda_{d}
\sum_{j \in \mbox{\boldmath $N$}(i)}
(1-2\,{\rm e}^{-\beta_{d}
\parallel d_{i}-\langle d_{j} \rangle_{\beta}^{\rm mf} \parallel^{2}}
)(1-\langle l(i,j) \rangle_{\beta}^{\rm mf})
\end{eqnarray*}
and obtain for $l(i,j)$ as 
\begin{eqnarray*}
&& E \simeq  E^{0}(l(i,j)) \nonumber \\
\mbox{} & \equiv &  
\lambda_{d}
(1-2\,{\rm e}^{-\beta_{d}
\parallel \langle d_{i} \rangle_{\beta}^{\rm mf} - \langle d_{j} \rangle_{\beta}^{\rm mf} \parallel^{2}})(1-l(i,j)) \nonumber \\
\mbox{} & + &  \lambda_{s}
(1-l(i,j))
(1-2\delta(\langle s_{i} \rangle_{\beta}^{\rm mf} 
-\langle s_{j} \rangle_{\beta}^{\rm mf})) +  
\frac{\alpha_{l}\, l(i,j)}
{(x_{i}^{\tau}-x_{j}^{\tau})^{2}} 
\end{eqnarray*}
where $\delta (\cdots)$ stands for a delta-function. 
By means of the above approximated cost functions, one 
obtains the following 
self-consistent equations for $\forall_{i,j \in S}$. 
\begin{eqnarray*}
\langle s_{i} \rangle_{\beta}^{\rm mf} 
& = & 
\frac{\sum_{s_{i}=0}^{1}
s_{i}\,
{\rm e}^{-\beta E^{0}(s_{i})}}
{\sum_{s_{i}=0}^{1}
{\rm e}^{-\beta E^{0}(s_{i})}} \nonumber \\
\mbox{} & \equiv &  
\Phi_{\beta}^{s}
(\langle d_{i} \rangle_{\beta}^{\rm mf}, 
\langle l(i,j) \rangle_{\beta}^{\rm mf}, 
\langle s_{i} \rangle_{\beta}^{\rm mf},\cdots) \\
\langle d_{i} \rangle_{\beta}^{\rm mf} 
& = & 
\frac{\sum_{d_{i}=0}^{d_{\rm max}-1}
d_{i}\,
{\rm e}^{-\beta E^{0}(d_{i})}}
{\sum_{d_{i}=0}^{d_{\rm max}-1}
{\rm e}^{-\beta E^{0}(d_{i})}} \nonumber \\
\mbox{} & \equiv & 
\Phi_{\beta}^{d}
(\langle s_{i} \rangle_{\beta}^{\rm mf}, 
\langle d_{j} \rangle_{\beta}^{\rm mf}, 
\langle l(i,j) \rangle_{\beta}^{\rm mf}, \cdots) \\
\langle l(i,j) \rangle_{\beta}^{\rm mf} 
& = & 
\frac{\sum_{l(i,j)=0}^{1}
l(i,j)\,
{\rm e}^{-\beta E^{0}(l(i,j))}}
{\sum_{l(i,j)=0}^{1}
{\rm e}^{-\beta E^{0}(l(i,j))}} \nonumber \\
\mbox{} & \equiv & 
\Phi_{\beta}^{l}
(\langle d_{i} \rangle_{\beta}^{\rm mf}, 
\langle d_{j} \rangle_{\beta}^{\rm mf}, 
\langle s_{i} \rangle_{\beta}^{\rm mf}, 
\langle s_{j} \rangle_{\beta}^{\rm mf},\cdots)
\end{eqnarray*}
Regarding the above 
self-consistent equations 
with respect to single-site averages  
as the following `non-linear maps':  
\begin{eqnarray}
&& \langle s_{i} \rangle_{\beta}^{{\rm mf} (t + 1)} 
=  
\Phi_{\beta}^{s}
(\langle d_{i} \rangle_{\beta}^{{\rm mf} (t)}, 
\langle l(i,j) \rangle_{\beta}^{{\rm mf} (t)}, 
\langle s_{i} \rangle_{\beta}^{{\rm mf} (t)}, \cdots) \nonumber \\
\label{eq:mf_s} \\
&& \langle d_{i} \rangle_{\beta}^{{\rm mf} (t+1)}  =  
\Phi_{\beta}^{d}
(\langle s_{i} \rangle_{\beta}^{{\rm mf} (t)}, 
\langle d_{j} \rangle_{\beta}^{{\rm mf} (t)}, 
\langle l(i,j) \rangle_{\beta}^{{\rm mf} (t)}, \cdots) \nonumber \\
\label{eq:mf_d} \\ 
&& \langle l(i,j) \rangle_{\beta}^{{\rm mf} (t+1)} \nonumber \\
\mbox{} & = &   
\Phi_{\beta}^{l}
(\langle d_{i} \rangle_{\beta}^{{\rm mf} (t)}, 
\langle d_{j} \rangle_{\beta}^{{\rm mf} (t)}, 
\langle s_{i} \rangle_{\beta}^{{\rm mf} (t)}, 
\langle s_{j} \rangle_{\beta}^{{\rm mf} (t)}, \cdots) 
\label{eq:mf_l}
\end{eqnarray}
we look for the steady states of the above maps 
which should satisfy  
the following convergence condition.
\begin{eqnarray}
\epsilon_{t} & \equiv &  
N^{-1}\{
\parallel\langle \mbox{\boldmath $s$} \rangle_{\beta}^{{\rm mf} (t)}
-\langle \mbox{\boldmath $s$} \rangle_{\beta}^{{\rm mf} (t -1)}\parallel^{2} \nonumber \\
\mbox{} & + &   \parallel\langle \mbox{\boldmath $d$} \rangle_{\beta}^{{\rm mf} (t)} 
 -  \langle \mbox{\boldmath $d$} \rangle_{\beta}^{{\rm mf} (t -1)}\parallel^{2} \nonumber \\
 \mbox{} & + &  
\parallel\langle \mbox{\boldmath $l$} \rangle_{\beta}^{{\rm mf} (t)}
-\langle \mbox{\boldmath $l$} \rangle_{\beta}^{{\rm mf} (t -1)} \parallel^{2}
\}^{1/2} < \epsilon 
\end{eqnarray}
where $\epsilon$ should be a small value, say $\epsilon=1.0 \times 10^{-5}$. 
In general, a control parameter $\beta$ is time-dependent 
variable as $\beta (t)$ and the MAP estimate is obtained 
by controlling it as $\beta (t) \to \infty$ as 
$t \to \infty$. 
On the other hand, the MPM estimate is constructed by setting 
the $\beta$ to $1$ during the above iterations.

Generally speaking, 
the steady state $\langle \cdots \rangle_{\beta}^{{\rm mf}(\infty)}$ is 
different from $\langle \cdots \rangle_{\beta}$ which is a 
solution of the self-consistent equations, however, 
it might assume that  the $\langle \cdots \rangle_{\beta}^{{\rm mf}(\infty)}$ 
more likely to be close to $\langle \cdots \rangle_{\beta}$ 
if the landscape of the cost is not so complicated like {\it spin glasses} \cite{SpinGlass}. 
\section{Preliminary : divergence of mean-fields}
\label{sec:PRE}
\begin{figure}[ht]
\begin{center}
\includegraphics[width=3cm]{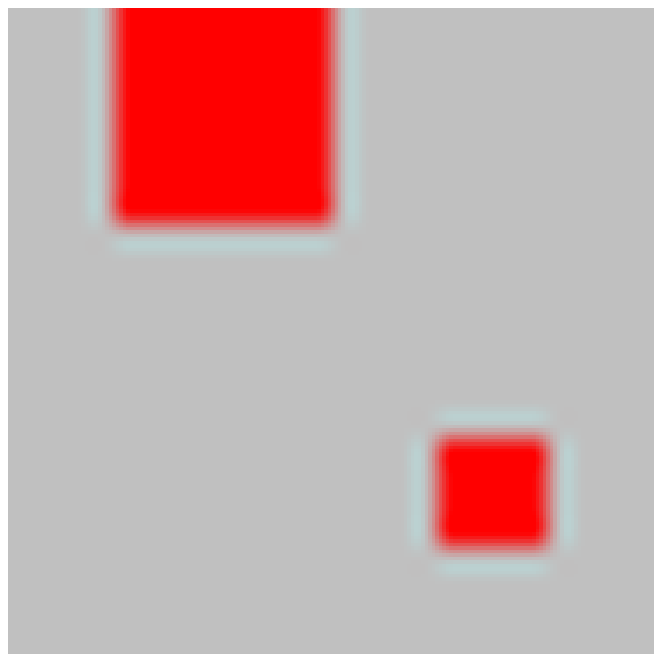} \hspace{0.2cm}
\includegraphics[width=3cm]{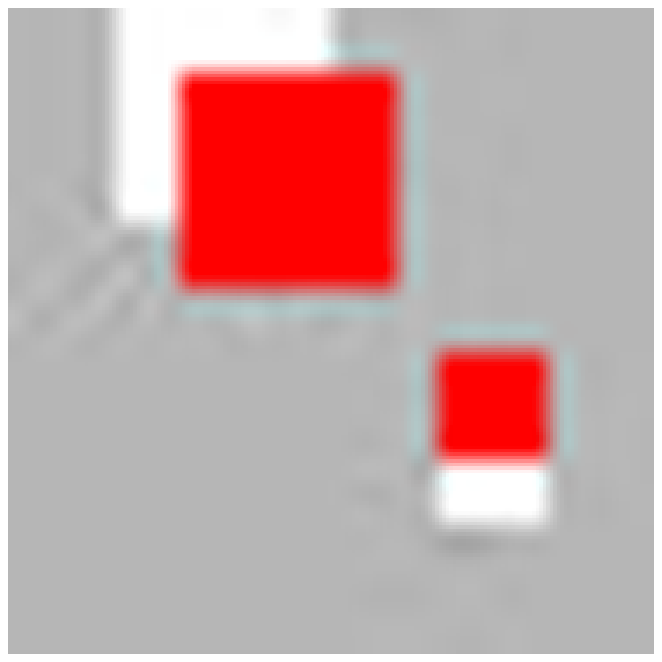} \\
\includegraphics[width=5cm]{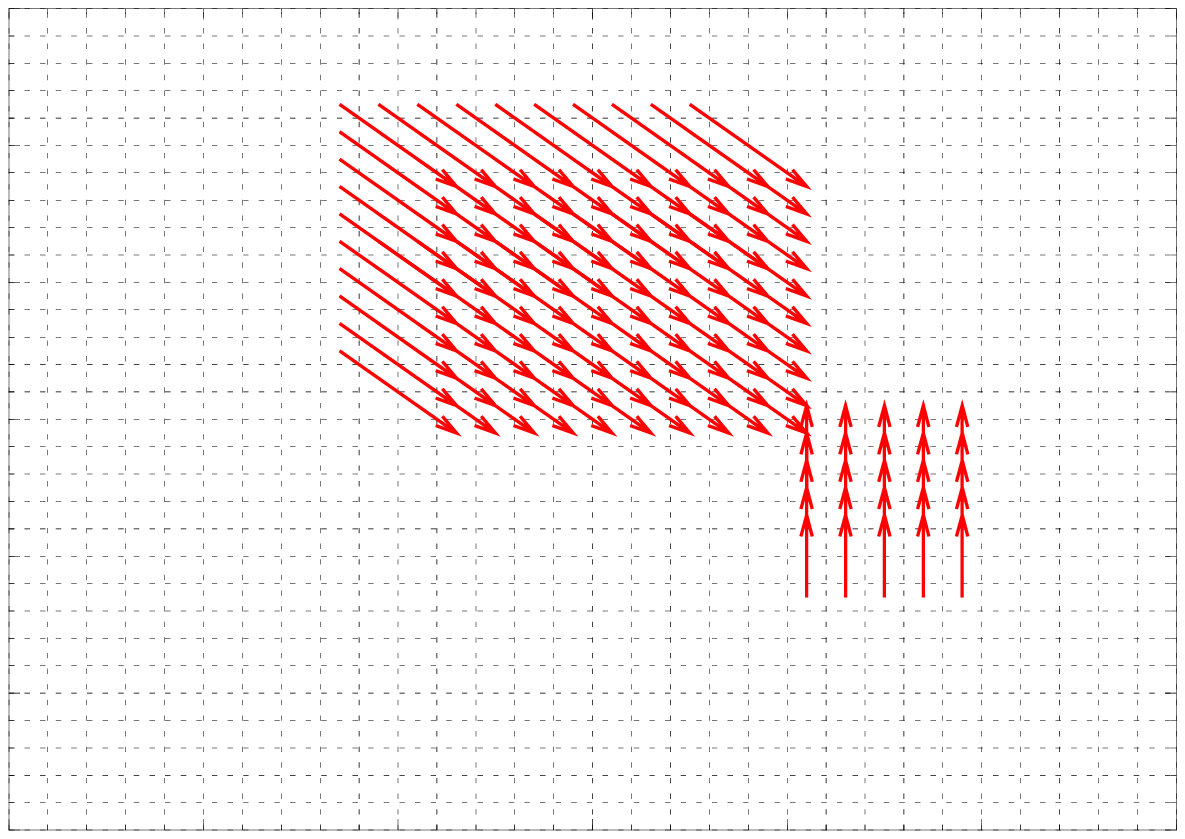}
\end{center}
\caption{
Typical artificial images as 
a set of successive two video-frames. 
Image before moving (upper left) 
and image after moving (upper right). 
The lower panel shows 
`true' motion velocity fields 
for the situation given by the upper 
panels. 
In the above images, 
arbitrary grayscales are given to 
the segmentation areas and the region 
in which the objects are located.  }
\label{fig:fg00}
\end{figure}

To check the usefulness of the above procedure, we examine 
our mean-field algorithm to 
infer the motion velocity fields for 
a given set of two successive frames 
shown in Fig. \ref{fig:fg00}.  
It should be noted that 
these two frames are artificially given and 
obviously, the true motion velocity vector fields 
are now explicitly provided for us to 
check the usefulness of our mean-field algorithm. 

Generally speaking in the Bayesian inference, 
setting the hyper-parameters 
appearing in the probabilistic 
model is one of the quite important tasks  
and here we examine the values 
$(\beta, 
\sigma^{2},
\lambda_{d}, \beta_{d}, 
\alpha_{l},
T_{s},\lambda_{s})=
(1,0.2,2.5,4,200,5,2)$ which were given ad-hoc by 
Zhang and Hanouer (1995). 
\begin{figure}[ht]
\begin{center}
\includegraphics[width=3.5cm,angle=90]{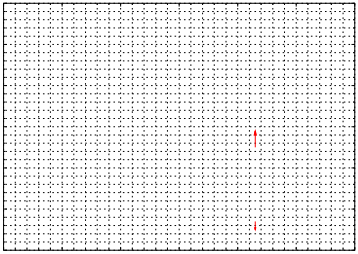}
\end{center}
\caption{ 
The resultant velocity fields 
calculated by the choice 
of hyper-parameters 
$(\beta, 
\sigma^{2},
\lambda_{d}, \beta_{d}, 
\alpha_{l},
T_{s},\lambda_{s})=
(1,0.2,2.5,4,200,5,2)$. 
The velocity fields shrink to 
a few points with small lengths. }
\label{fig:fg02}
\end{figure}
We find that for the above choice of 
the hyper-parameter causes a divergence of 
the mean-fields 
such as 
$\langle s_{i} \rangle_{\beta}^{\rm mf}$ 
due to the regularization terms 
$(1/2\sigma^{2})
(1-\langle s_{i} \rangle_{\beta}^{\rm mf})
(x_{i}^{\tau}-x_{i-d_{i}}^{\tau-1})^{2}$ or 
$-(s_{i}/2\sigma^{2})
(x_{i}^{\tau}-x_{i-\langle s_{i} \rangle_{\beta}^{\rm mf}}^{\tau-1})^{2}$ 
which appear in 
the mean-field equations. 
We show the 
resultant velocity fields 
calculated by the above choice 
of hyper-parameters in Fig. \ref{fig:fg02}. 
We find that the velocity fields shrink to 
a few points with small lengths
and one apparently fails to 
estimate the true velocity fields. 
\subsection{Optimization of scaling factor}
The origin of the above difficulty 
apparently comes from the divergence of these 
regularization terms 
evaluated for two extremely different values of 
pixels, for instance, say 
$x_{i}^{\tau}=255$ and 
$x_{i-d_{i}^{\tau}}^{\tau-1}=0$ which 
leads to ${\rm e}^{(255-0)^{2}} \sim \infty$. 
This fact tells us that 
there exist several serious cases 
(combinations of two consecutive video-frames) 
for which the ad-hoc 
hyper-parameter selection 
causes this type of divergence during the iteration of 
mean-field equations. 

To avoid the essential difficulty, we rescale the 
hyper-parameter $\sigma^{2}$ as 
$\sigma^{2} \mapsto \mu \sigma^{2}$ and 
optimizing the scaling factor $\mu$ from 
the view point of several different performance measures. 
\subsubsection{Performance measures}
We first introduce two different kinds of 
mean-square errors as 
average-case performance measures 
to determine the optimal scaling factor $\mu$. 
\begin{eqnarray}
D_{1} (\mu) & \equiv &  
\frac{1}{N_{1}}
\sum_{i=1}^{N} 
(1-s_{i}) \parallel d_{i}^{(0)}-d_{i} \parallel^{2} 
\label{eq:mse1} \\
D_{2} (\mu) & \equiv & 
\frac{1}{N_{2}}\sum_{i=1}^{N}
s_{i}\, \parallel d_{i}^{(0)}-d_{i} \parallel^{2}
\label{eq:mse2}
\end{eqnarray}
where 
$N_{1} \equiv \sum_{i=1}^{N}
(1-s_{i}), 
N_{2} \equiv \sum_{i=1}^{N} s_{i}$ 
and we should keep in mind that 
$N=N_{1}+N_{2}$ holds.  
$\mbox{\boldmath $d$}^{(0)}$ is 
a true velocity field for a given set of 
two successive images shown 
in Fig. \ref{fig:fg00}. 
Thus, 
the $D_{1}$ denotes 
the mean-square error defined by 
the difference between 
the true and the estimated velocity fields 
for zero segmentation regions. 
On the other hand, 
$D_{2}$ is 
the mean-square error evaluated for 
non-zero segmentation regions.

We also introduce the bit-error rate which is 
defined as the number of 
estimated pixels which are 
different from the true ones. 
Namely, we use 
\begin{eqnarray}
\delta_{1}(\mu)  & \equiv & 
\frac{1}{N_{1}}
\sum_{i=1}^{N}
(1-s_{i})\,
\hat{\delta}_{d_{i}^{0},d_{i}} \\
\delta_{2}(\mu) & \equiv &    
\frac{1}{N_{2}}
\sum_{i=1}^{N}
 s_{i}\,
\hat{\delta}_{d_{i}^{0},d_{i}}
\end{eqnarray}
where $\hat{\delta}_{x,y}$ means a Kronecker's delta which is defined by
\begin{eqnarray}
\hat{\delta}_{d_{i}^{0},d_{i}} & \equiv & 
\delta_{v_{x}^{0}(i),v_{x}(i)}
\delta_{v_{y}^{0}(i),v_{y}(i)}
\end{eqnarray} 
where $\delta_{x,y}$ is a `conventional' Kronecker's delta.
\begin{figure}[ht]
\begin{center}
\hspace{-0.6cm}
\includegraphics[width=4.5cm]{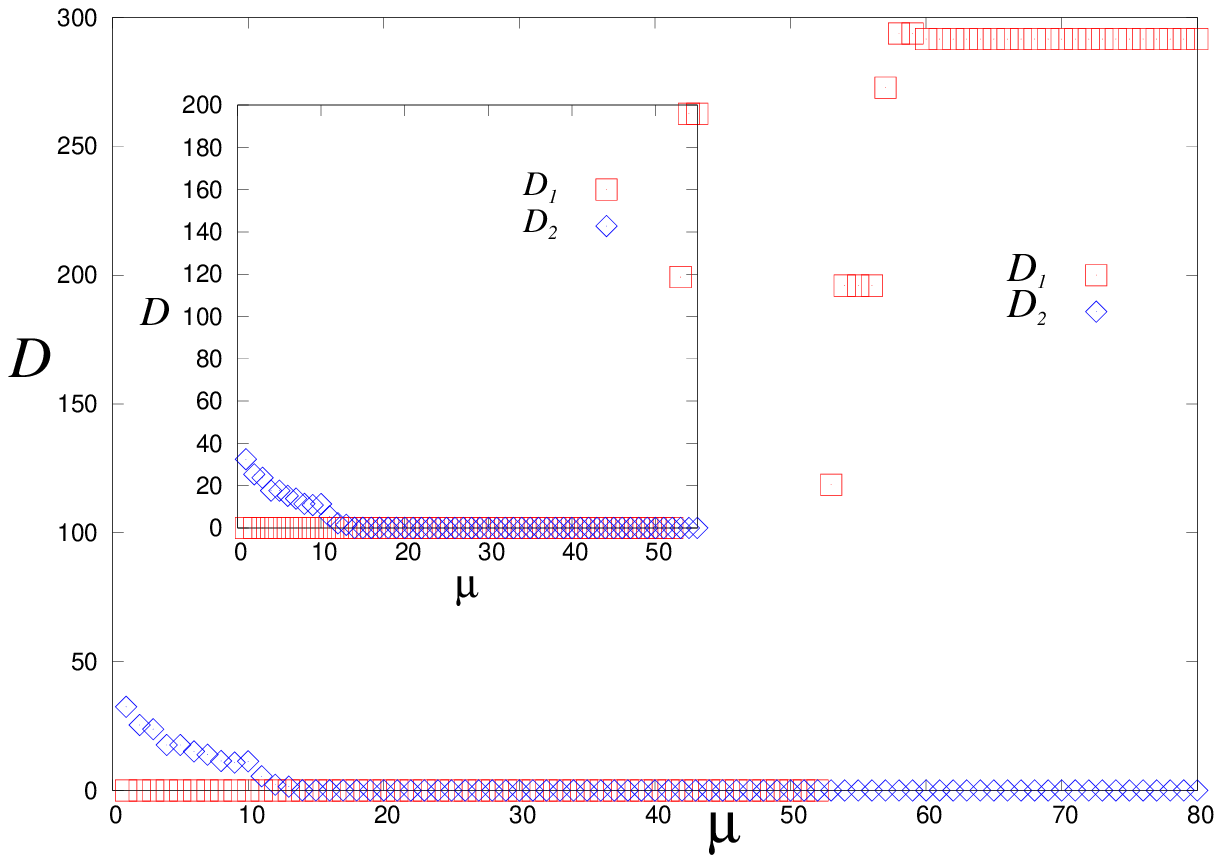} \hspace{-0.3cm}
\includegraphics[width=4.5cm]{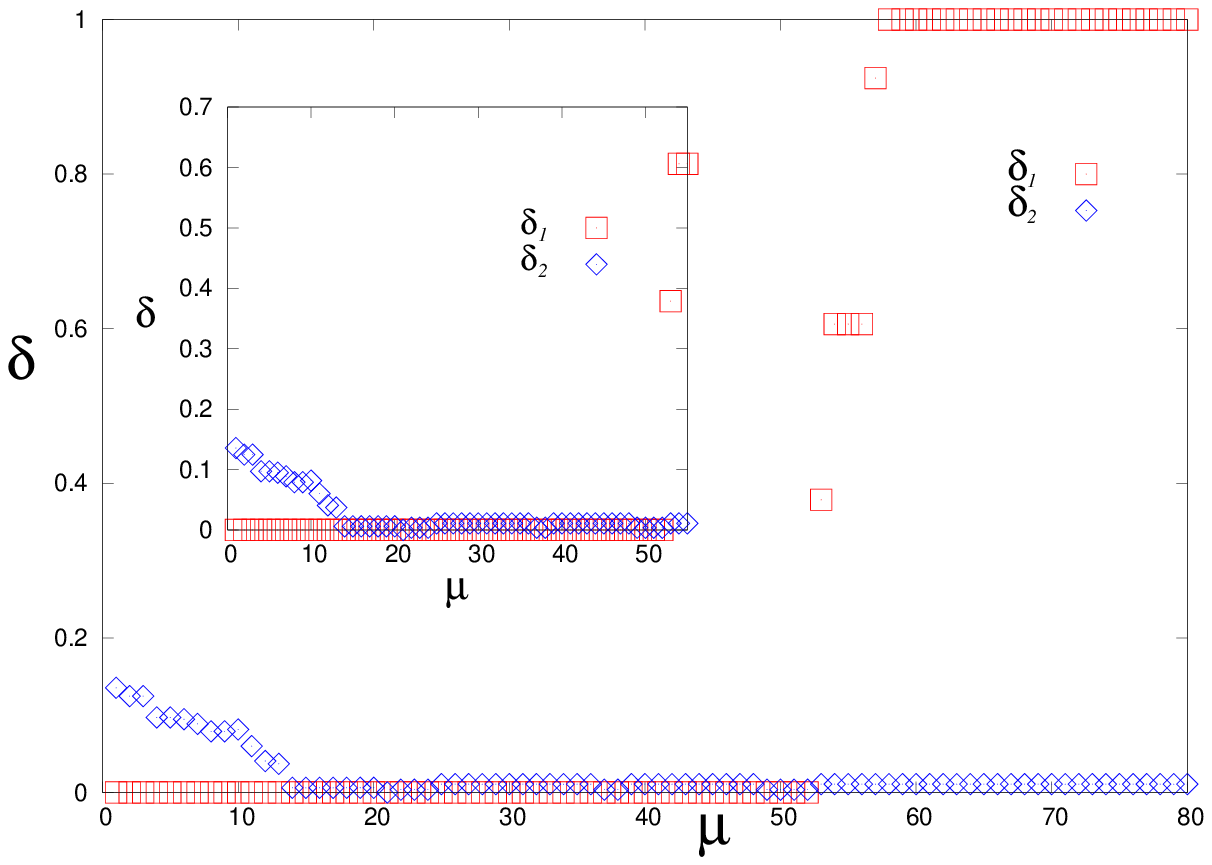} 
\includegraphics[width=5cm]{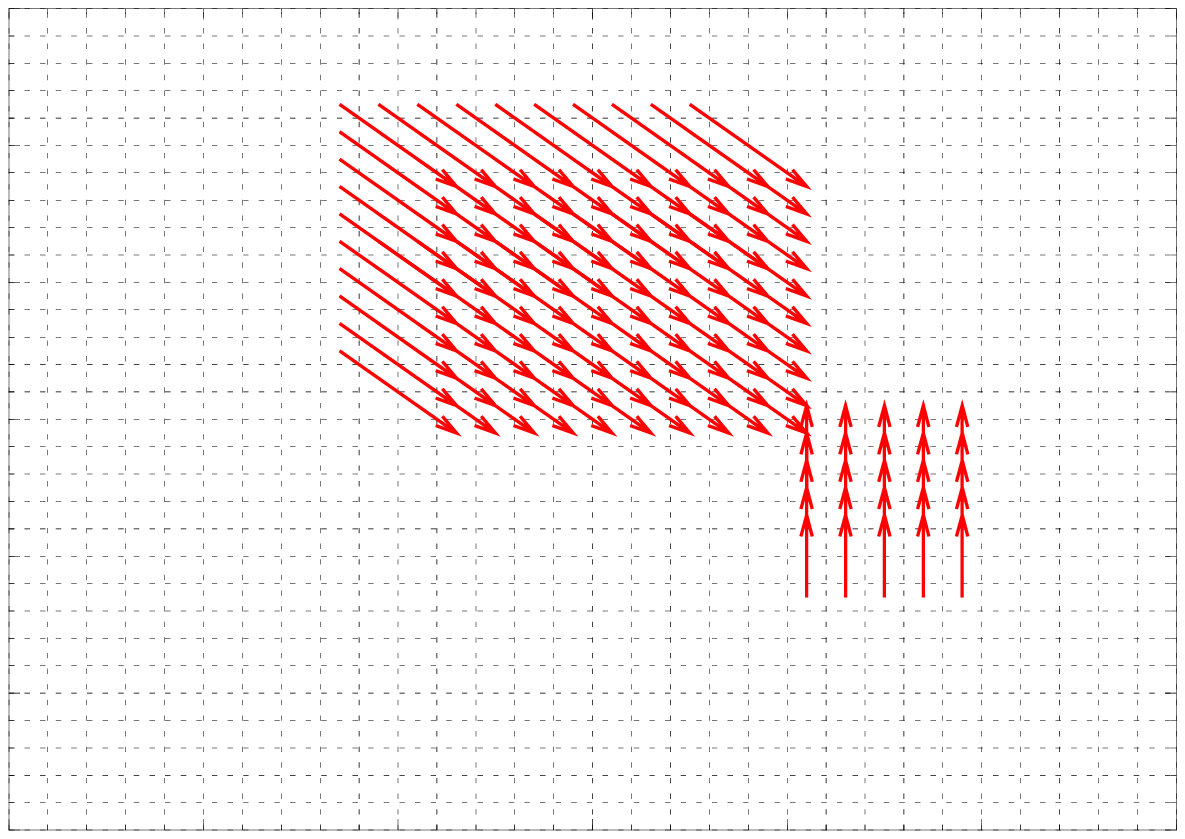} 
\end{center}
\caption{
Behaviour of two kinds of the mean-square errors 
$D_{1},D_{2}$ (upper left), 
the bit-error rates $\delta_{1},\delta_{2}$ 
(upper right) as a function of  scaling 
factor $\mu$. 
The lower panel shows 
the resultant velocity fields obtained by setting 
the optimal scaling factor 
$\mu_{*} \simeq 21$. 
The grayscale levels of the background and 
segmentation areas 
are $Q=0$ and 
$Q=40$, respectively. 
The grayscale levels for the 
moving object are 
distributed within the range $Q=10 \sim 30$. 
}
\label{fig:fg2}
\end{figure}
In Fig. \ref{fig:fg2}, 
we plot the behaviour of two kinds of the mean-square errors 
$D_{1},D_{2}$ (upper left), 
the bit-error rates $\delta_{1},\delta_{2}$ 
(upper right) as a function of  scaling 
factor $\mu$. 
The lower panel shows 
the resultant velocity fields obtained by setting 
the optimal scaling factor 
$\mu_{*} \simeq 21$. 
From these panels, we find that 
the resultant velocity fields are very close to 
the true fields when we set the 
scaling factor appropriately. 
However, 
the ad-hoc choice of the other 
hyper-parameters 
$(\beta, 
\sigma^{2},
\lambda_{d}, \beta_{d}, 
\alpha_{l},
T_{s},\lambda_{s})$
should not be confirmed for  
the best possible velocity fields estimation for a given other set 
of the successive images. 
To make matter worse, 
in practice, we can use 
neither mean-square error nor 
bit-error rate because 
these quantities require the information about the 
true fields $\mbox{\boldmath $d$}^{(0)}$ 
(for instance, see the definition of $D_{1}$). 
Therefore, we should seek some 
theoretical justifications to determine the 
optimal hyper-parameters. 
\section{Maximum marginal likelihood criteria}
\label{sec:ML}
In statistics, in order to determine 
the hyper-parameters 
$\mbox{\boldmath $\Xi$} \equiv 
\{\mu, \sigma, 
\lambda_{d},
\lambda_{s}, T_{s}, 
\alpha_{l}, \beta_{d}\}$ 
of the probabilistic model 
which contains 
latent variables 
$\mbox{\boldmath $\Sigma$} \equiv
\{\mbox{\boldmath $s$},
\mbox{\boldmath $d$},\mbox{\boldmath $l$}\}$, 
the so-called 
{\it maximum marginal likelihood 
estimation} is widely used. 
The marginal likelihood (the type-II likelihood) is defined by 
\begin{eqnarray}
-F_{\mbox{\boldmath $\Xi$}} (\mbox{\boldmath $x$}^{\tau},
\mbox{\boldmath $x$}^{\tau-1}) & \equiv & 
\log \sum_{\mbox{\boldmath $\Sigma$}}
P(\mbox{\boldmath $\Sigma$}| 
\mbox{\boldmath $x$}^{\tau},
\mbox{\boldmath $x$}^{\tau-1})
\label{eq:ML}
\end{eqnarray}
namely, the marginal 
likelihood 
is obtained by taking the sums of 
these latent variables in the (log) likelihood function.  
It should be noted that 
the above marginal likelihood is dependent on 
the `input' two successive frames 
$\mbox{\boldmath $x$}^{\tau},
\mbox{\boldmath $x$}^{\tau-1}$. 
We can easily show that 
the marginal likelihood is maximized at the 
true values of the hyper-parameters
$\mbox{\boldmath $\Xi$}^{0}$, 
namely,
\begin{eqnarray}
\left[
-F_{\mbox{\boldmath $\Xi$}^{0}}(\mbox{\boldmath $x$}^{\tau},
\mbox{\boldmath $x$}^{\tau-1}) 
\right]_{\mbox{\boldmath $x$}^{\tau},
\mbox{\boldmath $x$}^{\tau-1}} & \geq & 
\left[
-F_{\mbox{\boldmath $\Xi$}}
(\mbox{\boldmath $x$}^{\tau},
\mbox{\boldmath $x$}^{\tau-1})
\right]_{\mbox{\boldmath $x$}^{\tau},
\mbox{\boldmath $x$}^{\tau-1}}. \nonumber \\
\label{eq:marginalL}
\end{eqnarray}
where we defined 
the observable data-average by 
$\left[\cdots
\right] \equiv 
\sum_{\mbox{\boldmath $x$}^{\tau},
\mbox{\boldmath $x$}^{\tau-1}}
(\cdots) 
P_{\mbox{\boldmath $\Xi$}_{0}}
(\mbox{\boldmath $x$}^{\tau},
\mbox{\boldmath $x$}^{\tau-1})$. 
\subsection{Kullback-Leibler information}
Taking into account the fact that 
the Kullback-Leibler (KL) information can not be negative, 
we can easily show the inequality (\ref{eq:marginalL}).
 
Let us consider the KL information between 
the true probabilistic model 
$P_{\mbox{\boldmath $\Xi$}_{0}}
(\mbox{\boldmath $x$}^{\tau},
\mbox{\boldmath $x$}^{\tau-1})$ and 
the model  
$P_{\mbox{\boldmath $\Xi$}}
(\mbox{\boldmath $x$}^{\tau},
\mbox{\boldmath $x$}^{\tau-1})$. 
Then, from the definition of 
the KL information, we immediately have 
\begin{eqnarray}
&& KL(P_{\mbox{\boldmath $\Xi$}_{0}}||
P_{\mbox{\boldmath $\Xi$}}) \nonumber \\
\mbox{} & = & 
\sum_{\mbox{\boldmath $x$}^{\tau},
\mbox{\boldmath $x$}^{\tau-1}}
P_{\mbox{\boldmath $\Xi$}_{0}}
(\mbox{\boldmath $x$}^{\tau},
\mbox{\boldmath $x$}^{\tau-1})
\log 
\left\{
\frac{P_{\mbox{\boldmath $\Xi$}_{0}}
(\mbox{\boldmath $x$}^{\tau},
\mbox{\boldmath $x$}^{\tau-1})}
{P_{\mbox{\boldmath $\Xi$}}
(\mbox{\boldmath $x$}^{\tau},
\mbox{\boldmath $x$}^{\tau-1})}
\right\} \nonumber \\
\mbox{} & = & 
\sum_{\mbox{\boldmath $x$}^{\tau},
\mbox{\boldmath $x$}^{\tau-1}}
P_{\mbox{\boldmath $\Xi$}_{0}}
(\mbox{\boldmath $x$}^{\tau},
\mbox{\boldmath $x$}^{\tau-1})
\log 
P_{\mbox{\boldmath $\Xi$}_{0}}
(\mbox{\boldmath $x$}^{\tau},
\mbox{\boldmath $x$}^{\tau-1}) \nonumber \\
\mbox{} & - &  
\sum_{\mbox{\boldmath $x$}^{\tau},
\mbox{\boldmath $x$}^{\tau-1}}
P_{\mbox{\boldmath $\Xi$}_{0}}
(\mbox{\boldmath $x$}^{\tau},
\mbox{\boldmath $x$}^{\tau-1})
\log 
P_{\mbox{\boldmath $\Xi$}}
(\mbox{\boldmath $x$}^{\tau},
\mbox{\boldmath $x$}^{\tau-1}) \nonumber \\
\mbox{} & = & 
[-F_{\mbox{\boldmath $\Xi$}_{0}}
(\mbox{\boldmath $x$}^{\tau},
\mbox{\boldmath $x$}^{\tau-1})]_{\mbox{\boldmath $x$}^{\tau},
\mbox{\boldmath $x$}^{\tau-1}}-
[-F_{\mbox{\boldmath $\Xi$}}
(\mbox{\boldmath $x$}^{\tau},
\mbox{\boldmath $x$}^{\tau-1})]_{\mbox{\boldmath $x$}^{\tau},
\mbox{\boldmath $x$}^{\tau-1}} \nonumber \\
\mbox{} & \geq  & 0.
\end{eqnarray}
The equality holds if and only if $\mbox{\boldmath $\Xi$}=
\mbox{\boldmath $\Xi$}_{0}$. 
Therefore, 
the inequality (\ref{eq:marginalL}) holds and this means 
that the marginal likelihood takes its maximum at the 
true values of the hyper-parameters. 
We use this fact to determine 
the hyper-parameters. In other words, 
the marginal likelihood is regarded as 
a `cost function'  whose lowest energy states might be a 
candidate of the true hyper-parameters. 
\section{Hyper-parameter estimation}
As we saw in the previous section, 
we should determine hyper-parameters so 
as to minimize the marginal likelihood. 
In this section, we attempt to 
construct the Boltzmann-machine type 
learning equations 
which are derived by means of 
taking a gradient of the marginal likelihood 
with respect to the hyper-parameters. 
\subsection{Boltzmann-machine learning and its dynamics}
Let us define $\mbox{\boldmath $C$}(
\mbox{\boldmath $\Sigma$})$ 
as a conjugate statistics for 
the parameter 
$\mbox{\boldmath $\Xi$}$. 
Then, 
the Boltzmann-machine learning equation is obtained as 
\begin{eqnarray}
\frac{d \mbox{\boldmath $\Xi$}}
{dt} & = & -
\frac{\partial F_{\mbox{\boldmath $\Xi$}}
(\mbox{\boldmath $x$}^{\tau},
\mbox{\boldmath $x$}^{\tau-1})}
{\partial \mbox{\boldmath $\Xi$}} \nonumber \\
\mbox{} & = &  
-
\frac{\sum_{\mbox{\boldmath $\Sigma$}}
\mbox{\boldmath $C$}(
\mbox{\boldmath $\Sigma$})
P(\mbox{\boldmath $\Sigma$}|
\mbox{\boldmath $x$}^{\tau},
\mbox{\boldmath $x$}^{\tau+1})
}
{\sum_{\mbox{\boldmath $\Sigma$}}
P(\mbox{\boldmath $\Sigma$}|
\mbox{\boldmath $x$}^{\tau},
\mbox{\boldmath $x$}^{\tau+1})
}
\label{eq:koubai}
\end{eqnarray}
Namely, we have 
\begin{eqnarray}
\frac{dB}{dt}  & = &   
-\frac{\sum_{\mbox{\boldmath $\Sigma$}}
{\left\{\sum_{i}
(1-s_{i})(x_{i}^{\tau}
-x_{i-d_{i}}^{\tau-1})^{2}\right\}}{\rm e}^{-U}
}
{\sum_{\mbox{\boldmath $\Sigma$}}
{\rm e}^{-U}
}
\label{eq:C}  \\
\frac{d\lambda_{d}}
{dt} & = &  -    \frac{\sum_{\mbox{\boldmath $\Sigma$}}
\Lambda_{d}^{\beta_{d}}
(d_{i},d_{j},l(i,j)) 
{\rm e}^{-U}
}
{\sum_{\mbox{\boldmath $\Sigma$}}
{\rm e}^{-U}
}
\label{eq:lambda_d}  \\
\frac{d\lambda_{s}}
{dt} & = &  -\frac{\sum_{\mbox{\boldmath $\Sigma$}}
\Lambda_{d}^{\beta_{d}}
(d_{i},d_{j},l(i,j)){\rm e}^{-U}
}
{\sum_{\mbox{\boldmath $\Sigma$}}
{\rm e}^{-U}
}
\label{eq:lambda_s} \\
\frac{d\alpha_{l}}
{dt} & = &   -\frac{\sum_{\mbox{\boldmath $\Sigma$}}
{\left\{\sum_{i,j\in \mbox{\boldmath $N$}(i)}
\frac{l(i,j)}{(x_{i}^{\tau}-x_{j}^{\tau})^{2}}\right\}}{\rm e}^{-U}
}
{\sum_{\mbox{\boldmath $\Sigma$}}
{\rm e}^{-U}
}
\label{eq:A_l} \\
\frac{d\beta_{d}}
{dt}   & = &   -\frac{\sum_{\mbox{\boldmath $\Sigma$}}
{\cal B}_{d}^{\beta_{d}}
(l(i,j),d_{i},d_{j}) 
{\rm e}^{-U}
}
{\sum_{\mbox{\boldmath $\Sigma$}}
{\rm e}^{-U}
}
\label{eq:beta_d} \nonumber \\
\mbox{}\\ 
\frac{dT_{l}}
{dt}  &  = &  -\frac{\sum_{\mbox{\boldmath $\Sigma$}}
{\left\{\sum_{i}s_{i}\right\}}{\rm e}^{-U}
}
{\sum_{\mbox{\boldmath $\Sigma$}}
{\rm e}^{-U}
}
\label{eq:Ts}
\end{eqnarray}
where we defined 
\begin{eqnarray}
&& \hspace{-1.5cm} \Lambda_{d}^{\beta_{d}}
(d_{i},d_{j},l(i,j)) \nonumber \\
\mbox{} &  \equiv &  
\sum_{i,j\in \mbox{\boldmath $N$}(i)}
(1-2\,{\rm e}^{-\beta_{d} \parallel d_{i}-d_{j} \parallel^{2}})
(1-l(i,j)) \\
&& \hspace{-1.5cm}  {\cal B}_{d}^{\beta_{d}}
(l(i,j),d_{i},d_{j}) \nonumber \\
\mbox{} & \equiv &   
\sum_{i,j\in \mbox{\boldmath $N$}(i)}
(1-l(i,j)) \parallel d_{i}-d_{j} \parallel^{2}
{\rm e}^{-\beta_{d} \parallel d_{i}-d_{j}\parallel}
\end{eqnarray}
$B \equiv {1}/{2\mu \sigma^{2}}$ and 
$U \equiv \beta E(\mbox{\boldmath $\Sigma$}|
\mbox{\boldmath $x$}^{\tau},
\mbox{\boldmath $x$}^{\tau-1})$.
It should be noticed that 
the number of sums 
appearing in the right hand sides 
of the above equations comes up to exponential order and 
it is impossible for us to carry out them. 
\subsection{Hybridization of mean-field approximation and MCMC}
To overcome this computational difficulty, we 
utilize the mean-field approximation. 
We first replace the 
variables $\mbox{\boldmath $\Sigma$}$ 
with the corresponding expectations 
expect for the variables appearing in 
the brackets $\{\cdots\}$ in the right hand side 
of the learning equations. 
For instance,  $dB/dt= -\partial F_{\mbox{\boldmath $\Xi$}}/\partial B$ now leads to 
\begin{eqnarray}
\mbox{} \hspace{-1.5cm} && \frac{dB}
{dt} = -\frac{\sum_{s_{i},d_{i}}
{\left\{\sum_{i}
(1-s_{i})(x_{i}^{\tau}
-x_{i-d_{i}}^{\tau-1})^{2}\right\}}{\rm e}^{-\langle U \rangle_{s_{i},d_{i}}^{\rm mf}}
}
{\sum_{s_{i},d_{i}}
{\rm e}^{-\langle U \rangle_{s_{i},d_{i}}^{\rm mf}}
}
\label{eq:koubai_C2} 
\end{eqnarray}
\begin{eqnarray}
\mbox{}  && \hspace{-1.2cm} \langle U \rangle _{s_{i},d_{i}}^{\rm mf} \equiv  
B\sum_{i}
(1-s_{i})(x_{i}^{\tau}
-x_{i-d_{i}}^{\tau-1})^{2}  \nonumber \\
\mbox{} & + &   
\lambda_{d}\sum_{i,j\in \mbox{\boldmath $N$}(i)}
(1-2\,{\rm e}^{-\beta_{d} \parallel d_{i} -\langle d_{j} \rangle^{\rm mf} \parallel^{2}})
(1-\langle l(i,j) \rangle^{\rm mf}) \nonumber \\
\mbox{} & + & 
\lambda_{s}\sum_{i,j\in \mbox{\boldmath $N$}(i)}(1-\langle l(i,j) \rangle^{\rm mf})
(1-2\delta(s_{i}- \langle s_{j} \rangle^{\rm mf})) \nonumber \\
\mbox{} & + &   
\alpha_{l}
\sum_{i,j\in \mbox{\boldmath $N$}(i)}
\frac{\langle l(i,j) \rangle^{\rm mf}}{(x_{i}^{\tau}-x_{j}^{\tau})^{2}} + T_{s}
\sum_{i}s_{i} 
\label{eq:C2}
\end{eqnarray}
where we set $\beta=1$, namely, we calculate the MPM estimate in our framework. 
Using the same way as the above, 
$d\lambda_{d}/dt=-\partial F_{\mbox{\boldmath $\Xi$}}/\partial \lambda_{d}$ leads to 
\begin{eqnarray}
&& \hspace{-1.2cm} \frac{d\lambda_{d}}
{dt} =  -
\frac{\sum_{d_{i},d_{j},l_{ij}}
\Lambda_{d}^{\beta_{d}}(d_{i},d_{j},l(i,j)) 
{\rm e}^{-\langle U \rangle_{d_{i},d_{j},l_{ij}}^{\rm mf}}
}
{\sum_{d_{i},d_{j},l_{ij}}
{\rm e}^{-\langle U \rangle _{d_{i},d_{j},l_{ij}}^{\rm mf}}
}
\label{eq:koubai_d2}  \\
&& \hspace{-1.2cm} \Lambda_{d}^{\beta_{d}}(d_{i},d_{j},l(i,j)) \nonumber \\
\mbox{} & \equiv &  
\sum_{i,j\in \mbox{\boldmath $N$}(i)}
(1-2\,{\rm e}^{-\beta_{d} \parallel d_{i}-d_{j} \parallel^{2}})
(1-l(i,j)) \\
&& \hspace{-1.2cm} \langle U \rangle _{d_{i},d_{j},l_{ij}}^{\rm mf}  \equiv  
B\sum_{i}
(1-\langle s_{i} \rangle^{\rm mf})(x_{i}^{\tau}
-x_{i-d_{i}}^{\tau-1})^{2}  \nonumber \\
\mbox{} & + &   
\lambda_{d}\sum_{i,j\in \mbox{\boldmath $N$}(i)}
(1-2\,{\rm e}^{-\beta_{d} \parallel d_{i} - d_{j} \parallel^{2}})
(1-l(i,j)) \nonumber \\
\mbox{} & + & 
\lambda_{s}\sum_{i,j\in \mbox{\boldmath $N$}(i)}(1- l(i,j) )
(1-2\delta(\langle s_{i} \rangle^{\rm mf} - \langle s_{j} \rangle^{\rm mf})) \nonumber \\
\mbox{} & + &   
\alpha_{l}
\sum_{i,j\in \mbox{\boldmath $N$}(i)}
\frac{l(i,j) }{(x_{i}^{\tau}-x_{j}^{\tau})^{2}} + T_{s}
\sum_{i}\langle s_{i} \rangle^{\rm mf} 
\label{eq:lambda_d2}
\end{eqnarray}
The equations for the other parameters are also rewritten as 
\begin{eqnarray}
&& \hspace{-1.2cm} \frac{d\lambda_{s}}
{dt}  =  -
\frac{\sum_{l_{ij},s_{i},s_{j}}
\Lambda_{s}
(l(i,j),s_{i},s_{j}) 
{\rm e}^{-\langle U \rangle_{l_{ij},s_{i},s_{j}}^{\rm mf}}
}
{\sum_{l_{ij},s_{i},s_{j}}
{\rm e}^{-\langle U \rangle_{l_{ij},s_{i},s_{j}}^{\rm mf}}
}
\label{eq:koubai_s2} \\
&& \hspace{-1.2cm} \Lambda_{s}
(l(i,j),s_{i},s_{j}) \nonumber \\
\mbox{} & \equiv &   
\sum_{i,j\in \mbox{\boldmath $N$}(i)}(1-l(i,j))
(1-2\delta(s_{i}-s_{j})) \\
&& \mbox{}  \hspace{-1.2cm} \langle U \rangle _{s_{i},s_{j},l_{ij}}^{\rm mf} \equiv 
B\sum_{i}
(1-s_{i})(x_{i}^{\tau}
-x_{i-\langle d_{i} \rangle^{\rm mf}}^{\tau-1})^{2} \nonumber \\
\mbox{} & + &   
\lambda_{d}\sum_{i,j\in \mbox{\boldmath $N$}(i)}
(1-2\,{\rm e}^{-\beta_{d} \parallel \langle d_{i} \rangle^{\rm mf} 
-\langle d_{j} \rangle^{\rm mf} \parallel^{2}})
(1- l(i,j)) \nonumber \\
\mbox{} & + & 
\lambda_{s}\sum_{i,j\in \mbox{\boldmath $N$}(i)}(1-l(i,j))
(1-2\delta(s_{i}- s_{j})) \nonumber \\
\mbox{} & + &  
\alpha_{l}
\sum_{i,j\in \mbox{\boldmath $N$}(i)}
\frac{l(i,j)}{(x_{i}^{\tau}-x_{j}^{\tau})^{2}} + T_{s}
\sum_{i}s_{i} 
\label{eq:lambda_s2}
\end{eqnarray} 
\begin{eqnarray}
\frac{d \alpha_{l}}
{dt} & = & -
\frac{\sum_{l_{ij}}
{\left\{\sum_{i,j\in \mbox{\boldmath $N$}(i)}
\frac{l(i,j)}{(x_{i}^{\tau}-x_{j}^{\tau})^{2}}\right\}}
{\rm e}^{-\langle U \rangle_{l_{ij}}^{\rm mf}}
}
{\sum_{l_{ij}}
{\rm e}^{-\langle U \rangle_{l_{ij}}^{\rm mf}}
}
\label{eq:koubai_A_l2} \\
\mbox{}  \langle U \rangle _{l_{ij}}^{\rm mf}  & \equiv &
B\sum_{i}
(1-\langle s_{i} \rangle)(x_{i}^{\tau}
-x_{i-\langle d_{i} \rangle }^{\tau-1})^{2} \nonumber \\
\mbox{} & + &   
\lambda_{d}\sum_{i,j\in \mbox{\boldmath $N$}(i)}
(1-2\,{\rm e}^{-\beta_{d} \parallel \langle d_{i} \rangle^{\rm mf} 
-\langle d_{j} \rangle^{\rm mf} \parallel^{2}}) \nonumber \\
\mbox{} & \times & 
(1- l(i,j)) \nonumber \\
\mbox{} & + & 
\lambda_{s}\sum_{i,j\in \mbox{\boldmath $N$}(i)}(1-l(i,j)) \nonumber \\
\mbox{} & \times & 
(1-2\delta(\langle s_{i} \rangle^{\rm mf} - \langle s_{j} \rangle^{\rm mf})) \nonumber \\
\mbox{} & + & 
\alpha_{l}
\sum_{i,j\in \mbox{\boldmath $N$}(i)}
\frac{l(i,j)}{(x_{i}^{\tau}-x_{j}^{\tau})^{2}} + T_{s}
\sum_{i}\langle s_{i} \rangle^{\rm mf}
\label{eq:A_l2}
\end{eqnarray}
\begin{eqnarray}
&& \hspace{-0.5cm} \frac{d\beta_{d}}
{dt}  =  -
\frac{\sum_{d_{i},d_{j},l_{ij}}
{\cal B}_{d}^{\beta_{d}}(l(i,j),d_{i},d_{j}) 
{\rm e}^{-\langle U \rangle_{d_{i},d_{j},l_{ij}}^{\rm mf}}
}
{\sum_{d_{i},d_{j},l_{ij}}
{\rm e}^{-\langle U \rangle_{d_{i},d_{j},l_{ij}}^{\rm mf}}
}
\label{eq:koubai_beta_d2} \\
&&  \hspace{-0.8cm} {\cal B}_{d}^{\beta_{d}}(l(i,j),d_{i},d_{j})  \nonumber \\
\mbox{} & \equiv &   \sum_{i,j\in \mbox{\boldmath $N$}(i)}
(1-l(i,j)) \parallel d_{i}-d_{j} \parallel^{2}
{\rm e}^{-\beta_{d} \parallel d_{i}-d_{j} \parallel} \\
&&  \langle U \rangle _{d_{i},d_{j},l_{ij}}^{\rm mf}   \equiv 
B\sum_{i}
(1-\langle s_{i} \rangle^{\rm mf})(x_{i}^{\tau}
-x_{i-d_{i}}^{\tau-1})^{2} \nonumber \\
\mbox{} & + &   
\lambda_{d}\sum_{i,j\in \mbox{\boldmath $N$}(i)}
(1-2\,{\rm e}^{-\beta_{d} \parallel d_{i} -d_{j} \parallel^{2}})
(1-l(i,j)) \nonumber \\
\mbox{} & + & 
\lambda_{s}\sum_{i,j\in \mbox{\boldmath $N$}(i)}(1-l(i,j))
(1-2\delta(\langle s_{i} \rangle^{\rm mf} - \langle s_{j} \rangle^{\rm mf})) \nonumber \\
\mbox{} & + &  
\alpha_{l}
\sum_{i,j\in \mbox{\boldmath $N$}(i)}
\frac{l(i,j)}{(x_{i}^{\tau}-x_{j}^{\tau})^{2}} + T_{s}
\sum_{i}\langle s_{i} \rangle^{\rm mf} 
\label{eq:beta_d2} 
\end{eqnarray}
\begin{eqnarray}
\frac{dT_{s}}
{dt} & = & -
\frac{\sum_{s_{i}}
{\left\{\sum_{i}s_{i}\right\}}
{\rm e}^{-\langle U \rangle_{s_{i}}^{\rm mf}}
}
{\sum_{s_{i}}
{\rm e}^{-\langle U \rangle_{s_{i}}^{\rm mf}}
}
\label{eq:koubai_Ts2} \\
\mbox{}  \langle U \rangle _{s_{i}}^{\rm mf} 
 & \equiv & B\sum_{i}
(1-s_{i})(x_{i}^{\tau}
-x_{i-\langle d_{i} \rangle^{\rm mf} }^{\tau-1})^{2} \nonumber \\
\mbox{} & + &   
\lambda_{d}\sum_{i,j\in \mbox{\boldmath $N$}(i)}
(1-2\,{\rm e}^{-\beta_{d}|\langle d_{i} \rangle^{\rm mf} 
-\langle d_{j} \rangle^{\rm mf}|^{2}}) \nonumber \\
\mbox{} & \times &  
(1-\langle l(i,j) \rangle^{\rm mf}) \nonumber \\
\mbox{} & + & 
\lambda_{s}\sum_{i,j\in \mbox{\boldmath $N$}(i)} \nonumber \\
\mbox{} & \times &  (1-\langle l(i,j) \rangle^{\rm mf})
(1-2\delta(s_{i}- \langle s_{j} \rangle^{\rm mf})) \nonumber \\
\mbox{} & + &  
\alpha_{l}
\sum_{i,j\in \mbox{\boldmath $N$}(i)}
\frac{\langle l(i,j) \rangle^{\rm mf}}
{(x_{i}^{\tau}-x_{j}^{\tau})^{2}} + T_{s}
\sum_{i}s_{i} 
\label{eq:Ts2}
\end{eqnarray}
where 
$\langle \cdots \rangle^{\rm mf}$  
denotes a solution 
for the corresponding mean-field equation for a
given hyper-parameter set at time $t$ of the above learning equations : $\mbox{\boldmath $\Sigma$}^{(t)}$.   
There still exist several (it is still hard for us to treat by hand) sums in the above learning equations and 
it might be possible for us evaluate the sums also by the expectations 
in terms of mean-field approximation. 
However, for such treatment, the learning equations 
looks for the hyper-parameters which  
minimize the cost function instead of the `negative' marginal likelihood. 
From the view point of statistical physics, 
the marginal likelihood corresponds to 
the negative free energy and 
the mean-field treatment eliminates the entropy term. 
Therefore, if we rewrite the marginal likelihood by means of 
mean-field approximation, 
one obtains the negative cost function instead of the marginal likelihood. 
This means that we can not obtain appropriate 
 hyper-parameters in terms of the maximum marginal likelihood criteria. 
 For this reason, here 
 we use the Markov chain Monte Carlo method (MCMC) 
 to evaluate the sums appearing in the right hand sides of the learning equations. 
 
 In order to implement the learning equations 
 in computer, we discretize the derivative with respect to time $t$ 
 by means of Euler method such as 
 \begin{eqnarray*}
&& B(t+\Delta t)  =   B(t) \nonumber \\
\mbox{} & + &  \Delta t
\left\{\frac{\sum_{d_{i},s_{i}}
{\left\{\sum_{i}
(1-s_{i})(x_{i}^{\tau}
-x_{i-d_{i}}^{\tau-1})^{2}\right\}}e^{-\langle U \rangle_{s_{i},d_{i}}^{\rm mf}}
}
{\sum_{d_{i},s_{i}}
e^{-\langle U \rangle_{s_{i},d_{i}}^{\rm mf}}
}\right\}.
\end{eqnarray*}
Thus, we set the initial values of hyper-parameters 
to $\mbox{\boldmath $\Sigma$}^{(0)}$ and 
solve the mean-field equations. 
Then, we insert the solutions into the right hand sides of the above learning 
equations and evaluate the sums such as $\sum_{s_{i}}(\cdots)$ by 
the MCMC. 
After that, we update the hyper-parameters by the discretized learning equations 
and also update the time (step) as $t \mapsto t+1$. 
We repeat these procedures until 
each hyper-parameter converges to some finite value. 
Here we set $\Delta t=0.001$. 
The initial values $\mbox{\boldmath $\Sigma$}^{(0)}$ are the same values 
as those by Zhang and Hanouer (1995). 
\begin{figure}
\begin{center}
\includegraphics[width=4.4cm]{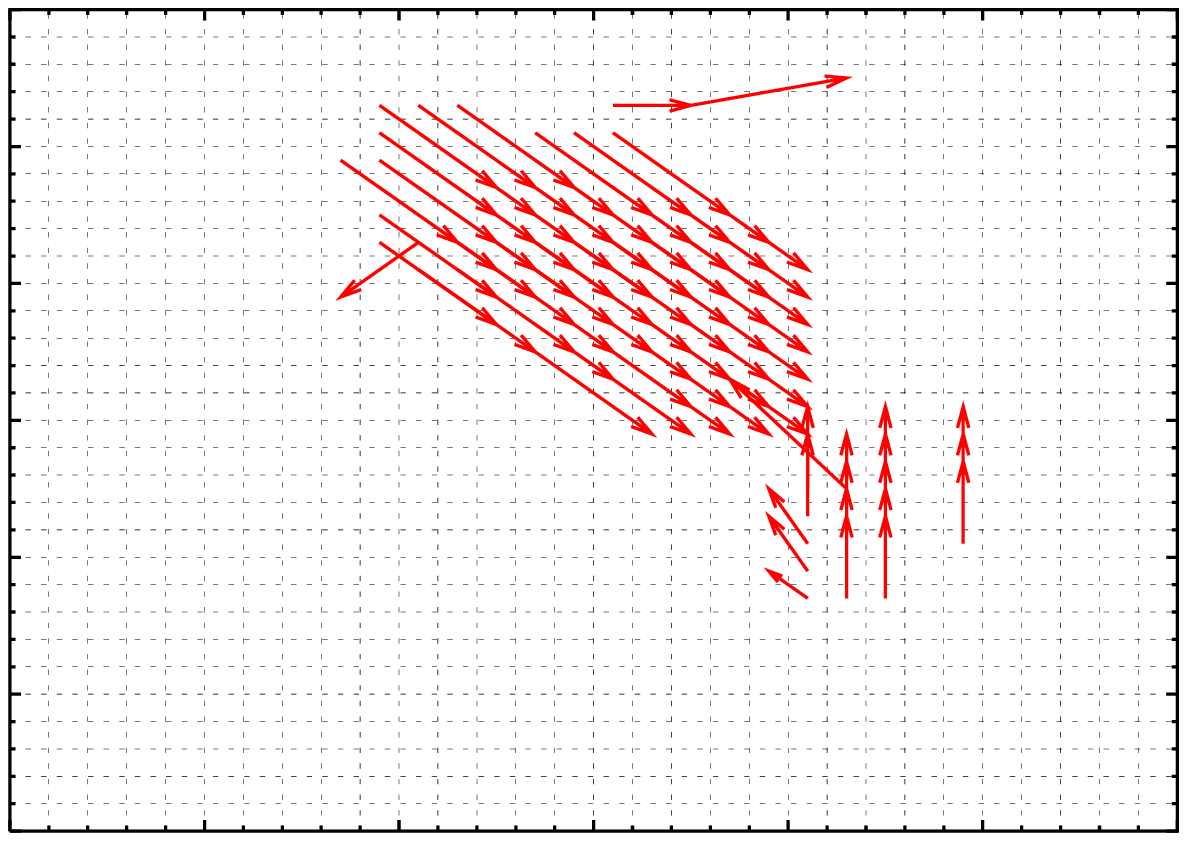} \hspace{-0.3cm}
\includegraphics[width=4.4cm]{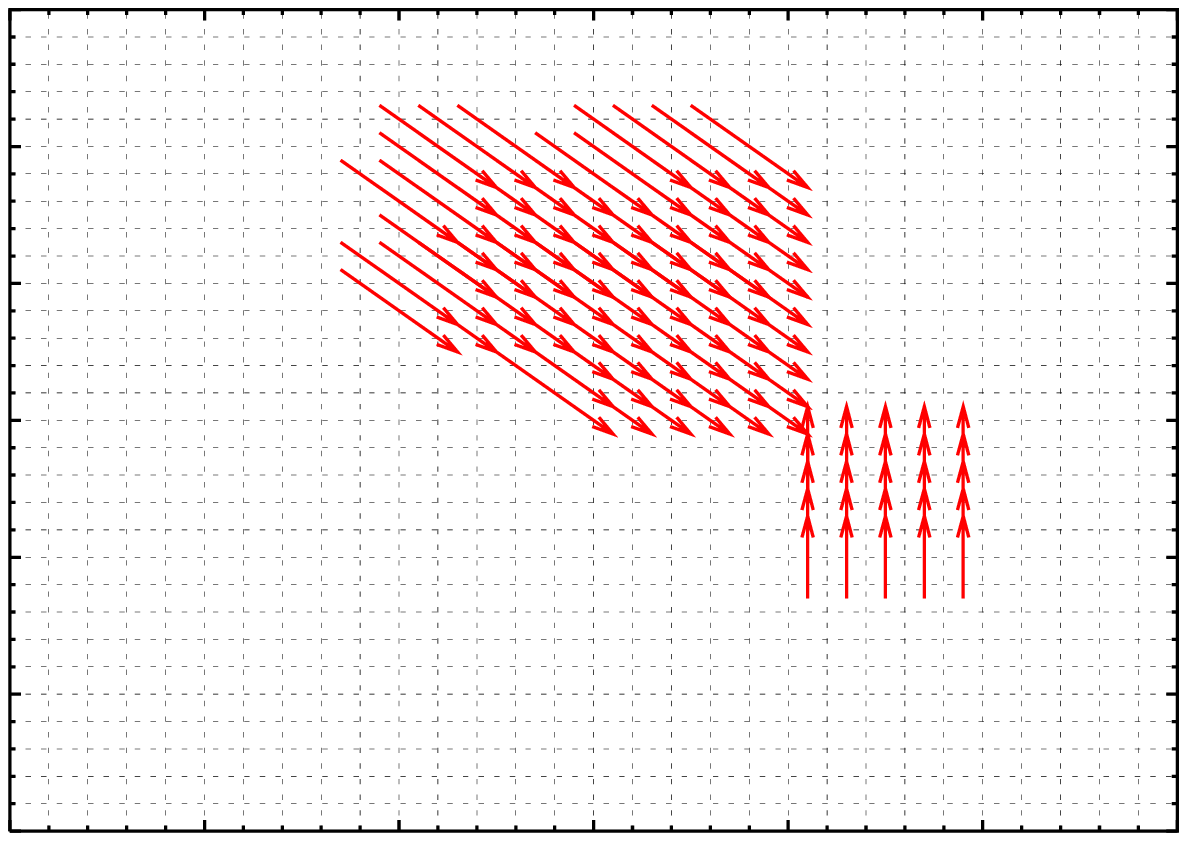}\\
\includegraphics[width=4.4cm]{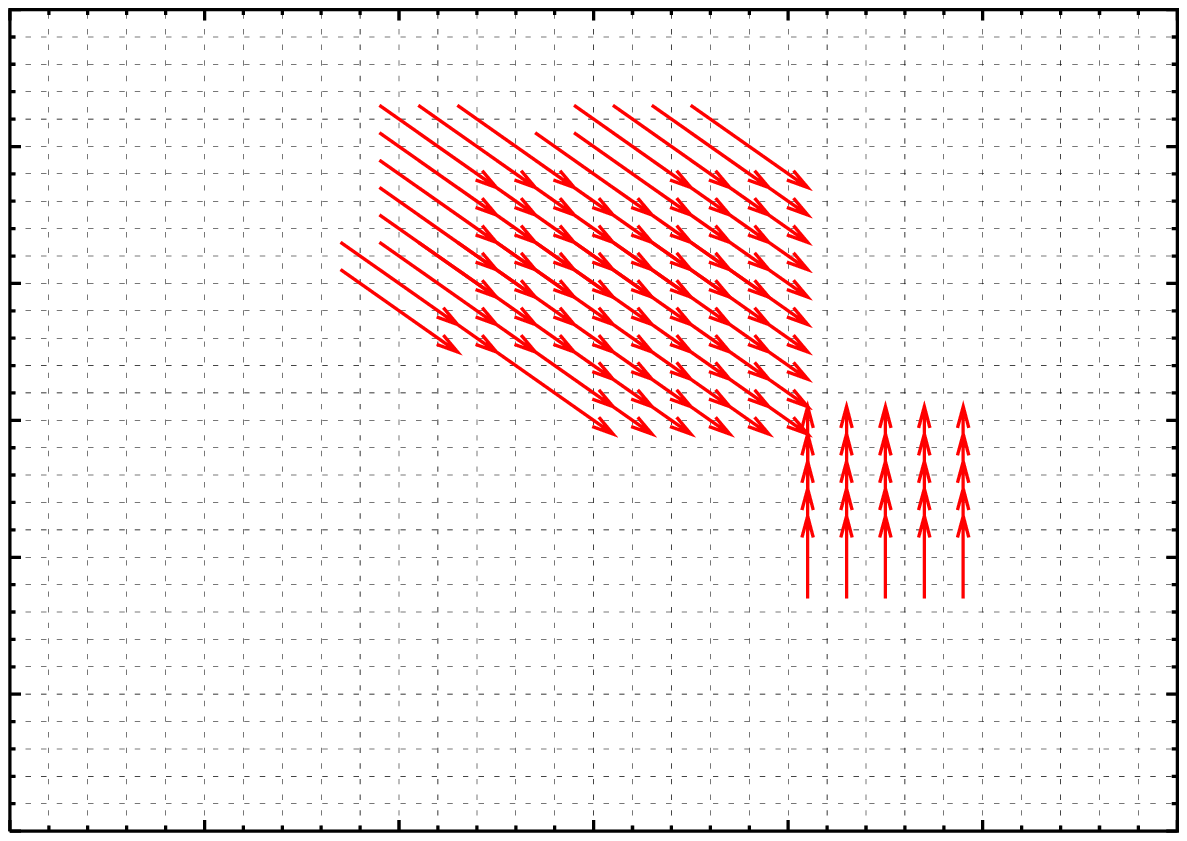} \hspace{-0.3cm}
\includegraphics[width=4.4cm]{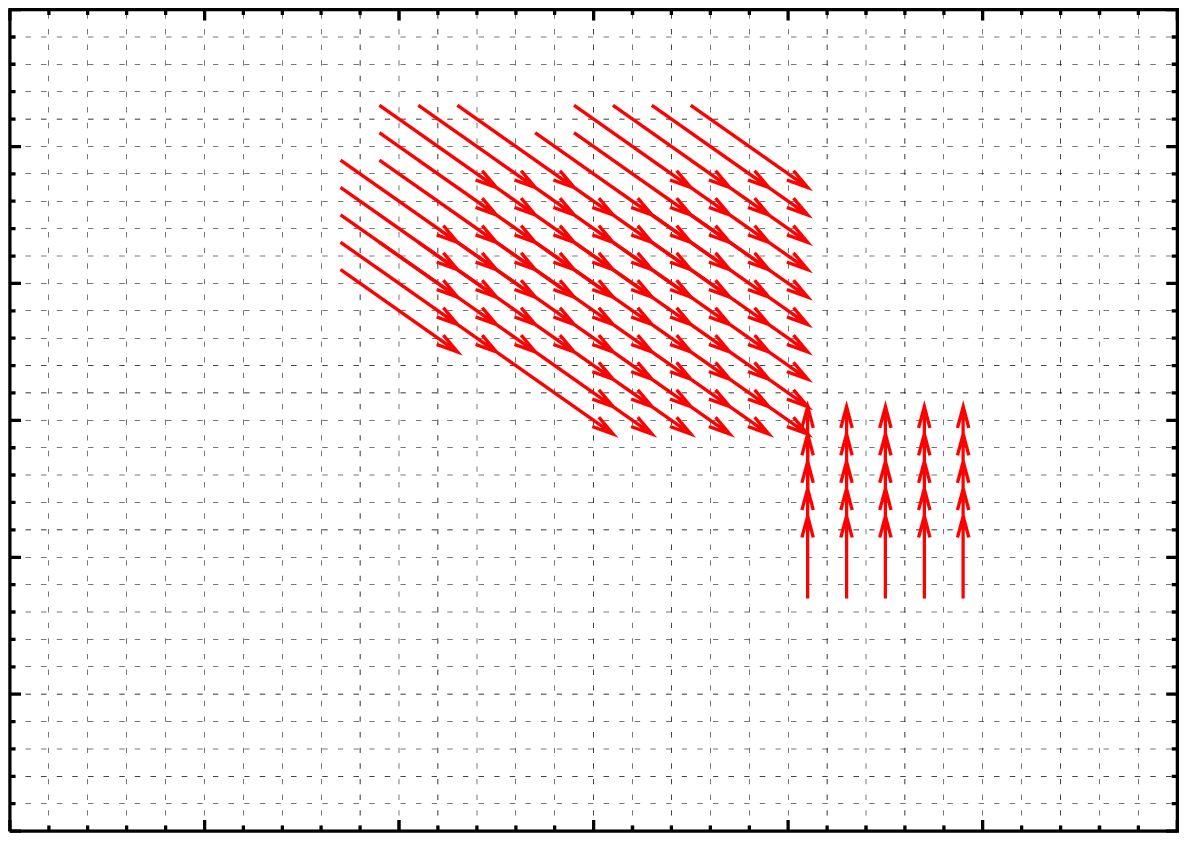}
\end{center}
\caption{\footnotesize 
Typical snapshots of velocity fields 
obtained 
by the method of 
hybridization of mean-field approximation and 
MCMC 
at time 
$t=0$ (upper left)，
$t=10$ (upper right), 
$t=20$ (lower left), 
$t=30$ (lower right)，
The case of 
$t=0$ corresponds to 
the result by 
Zhang and Hanouer (1995). }
\label{fig:jikken6-1_1}
\end{figure}
\mbox{}

In Fig. \ref{fig:jikken6-1_1}, we show 
the typical snapshots of velocity fields obtained 
by the method of 
hybridization of mean-field approximation and 
MCMC at time 
$t=0$ (upper left)，
$t=10$ (upper right), 
$t=20$ (lower left), 
$t=30$ (lower right)，
The case of 
$t=0$ corresponds to 
the result by 
Zhang and Hanouer (1995). 
From these panels, we find that 
our approach remarkably improves 
the performance of Zhang and Hanouer (1995).
\subsubsection{Average-case performance measures}
To evaluate the average-case performance 
more quantitatively, we 
introduce the following two kinds of 
performance measures. 
The first one is defined by 
\begin{eqnarray}
K & \equiv & \frac{1}{N}\sum_{i}(1-\cos\theta_{i})
\end{eqnarray}
where 
$\theta_{i}$ denotes an angle between 
the true velocity vector fields  
$\mbox{\boldmath $d$}^{0}=\{\vec{d}_{1}^{0},\cdots,\vec{d}_{N}^{0}\}$ 
and 
the estimated fields 
$\mbox{\boldmath $d$}=\{\vec{d}_{1},\cdots,\vec{d}_{N}\}$, 
that is explicitly given by 
$\cos \theta_{i} = \vec{d}_{i}^{0} 
\cdot \vec{d}_{i}/
\parallel \vec{d}_{i}^{0}\parallel  
\parallel \vec{d}_{i} \parallel$. 
From the above definition, the $K$ 
measures the error concerning mismatch of 
the direction of the estimated vector. 

Besides of the above $K$, we next introduce 
\begin{eqnarray}
L & \equiv & \frac{1}{N}
\sum_{i}\left(1
-\frac{\parallel \vec{d}_{i} \parallel}
{\parallel \vec{d}_{i}^{0}\parallel}
\right)
\end{eqnarray}
which measures the error concerning 
mismatch of the length of the estimated vector.
\begin{figure}[ht]
\begin{center}
\includegraphics[width=5.5cm,angle=270]{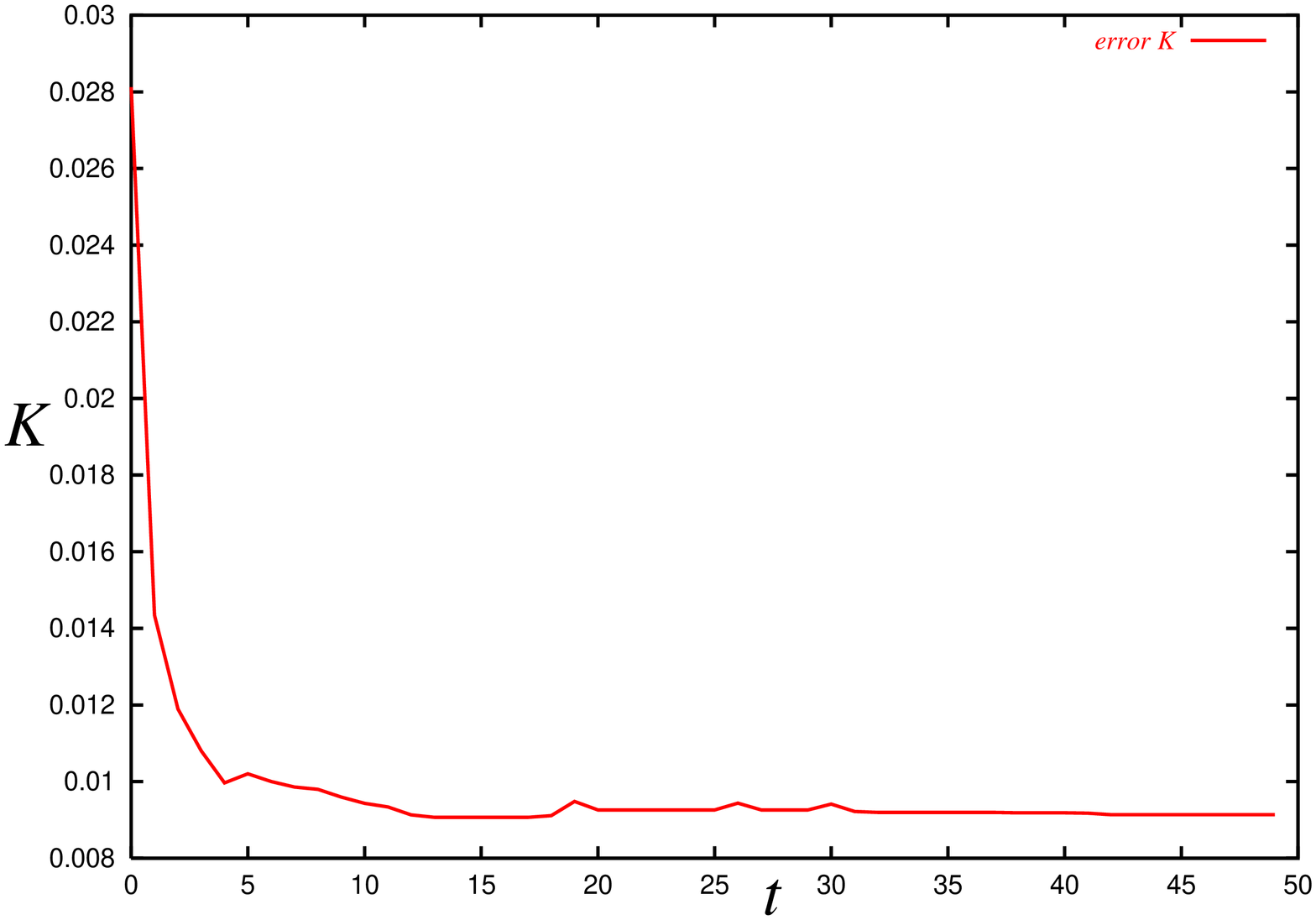} \\
\includegraphics[width=5.5cm,angle=270]{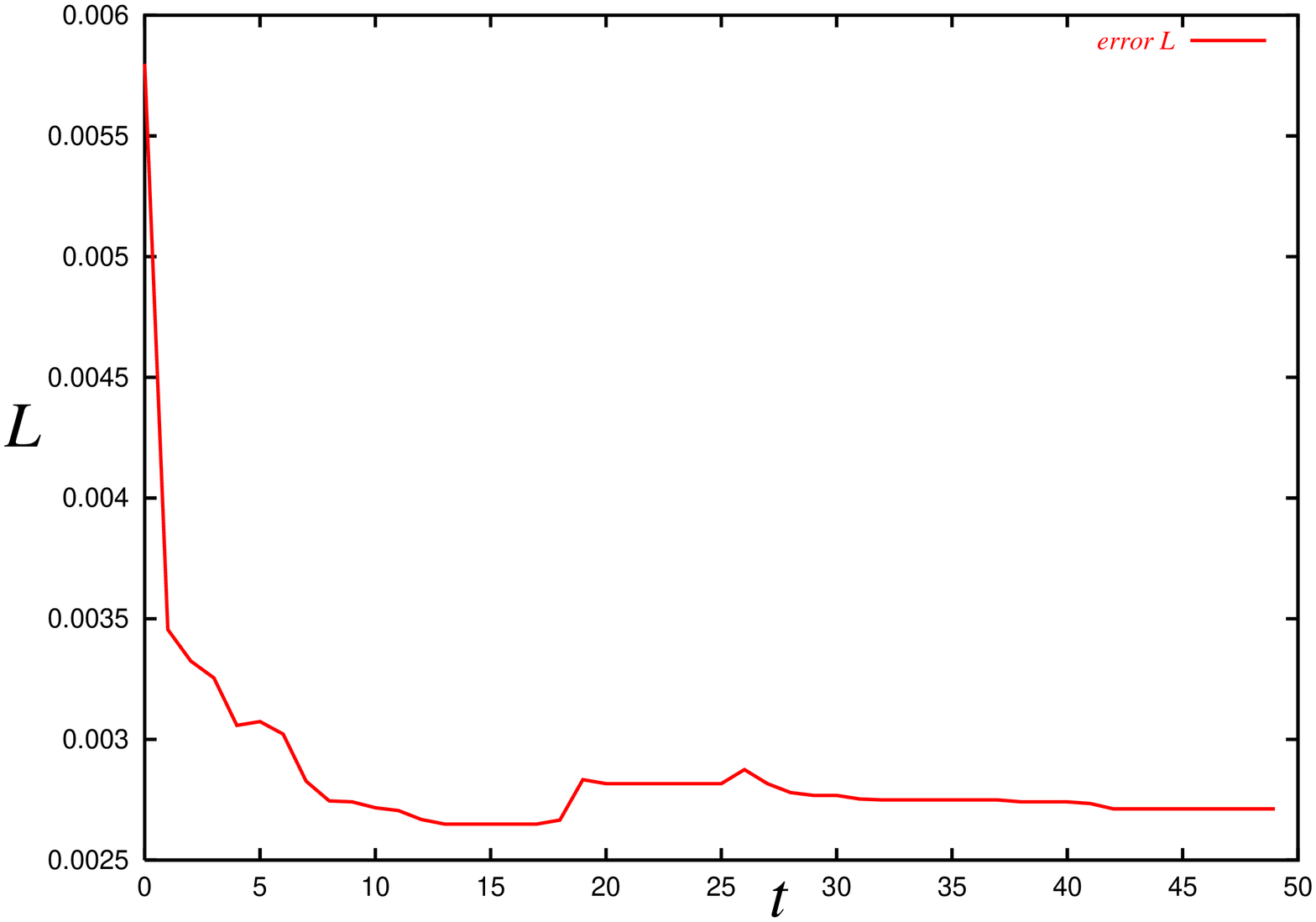} 
\end{center}
\caption{
Time dependence of the performance measures $K$ (upper panel) and 
$L$ (lower panel). We plot the average values of 
$K$ and $L$ over $20$-independent runs for various different choices of 
the successive two video-frames.}
\label{fig:jikken6-1_3}
\end{figure}
We show the results in Fig. \ref{fig:jikken6-1_3}. 
We plot the average values of 
$K$ and $L$ over $20$-independent runs for various different choices of 
the successive two video-frames.
From these two panels, we find that 
these two errors decreases monotonically on average 
during the proposed learning procedures. 
\subsubsection{Computational cost measure}
\begin{figure}[ht]
\begin{center}
\includegraphics[width=5.5cm,angle=270]{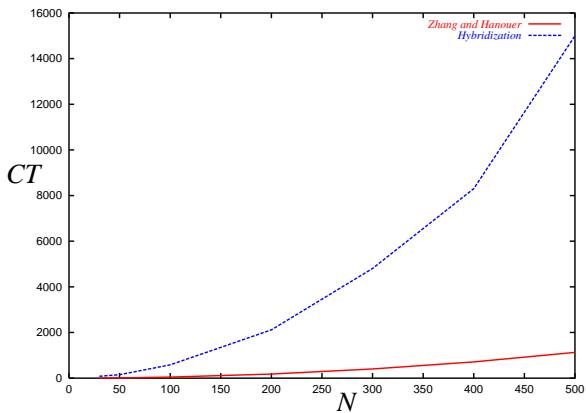} 
\end{center}
\caption{
Computational time (real CPU time) $CT$ [sec] until 
the algorithm converges as a function of system size $N$.}
\label{fig:jikken6-1_4}
\end{figure}

We next evaluate the computational cost. 
Obviously, our procedure requires 
us to take much longer time in comparison with 
the result by Zhang and Hanouer (1995) 
to obtain the results  because for each 
Euler step, one needs 
to solve the mean-field equations and 
one should carry out the MCMC at the same time. 
In Fig. \ref{fig:jikken6-1_4}, we plot the 
CPU time $CT$ [sec] as a function of system size $N$. 
The CPU time is measured 
in our PC ({\it DELL Optiplex960DT7, 
Core2QuadQ9400 2.66 GHz}).  
In the case of 
Zhang and Hanouer (1995), 
we measure the $CT$ [sec] as CPU time to 
proceed $50$-times mean-field iterations, whereas, 
in the case of our proposed procedure, 
the $CT$ is defined as CPU time 
to take $t=50$ in learning equations (for each of $t$,  
$50$-times mean-field iterations and $100$ Monte Carlo step are done). 
From 
Fig. \ref{fig:jikken6-1_4}, 
we find that the difference 
between two procedures 
increases exponentially, however, 
this fact does not mean that 
our proposed procedure 
is computationally inferior to the 
ad-hoc choice by Zhang and Hanouer (1995) 
because they found the value by `try and error'  manner 
and it might take a quite long time 
to determine the value although they did not 
mention this point explicitly in their paper. 
\subsection{Simple MCMC approach}
In general, the preciseness of the 
mean-field approximation is not so good. 
Here we attempt to use simple MCMC 
instead of hybridization of mean-field approximation and MCMC 
to calculate the expectations of quantities appearing in the learning 
equations over the 
posterior. Then, we compare the results with 
those obtained by hybridization of 
the mean-field approximation and 
the MCMC discussed in the previous subsection. 

\begin{figure}
\begin{center}
\includegraphics[width=4.4cm]{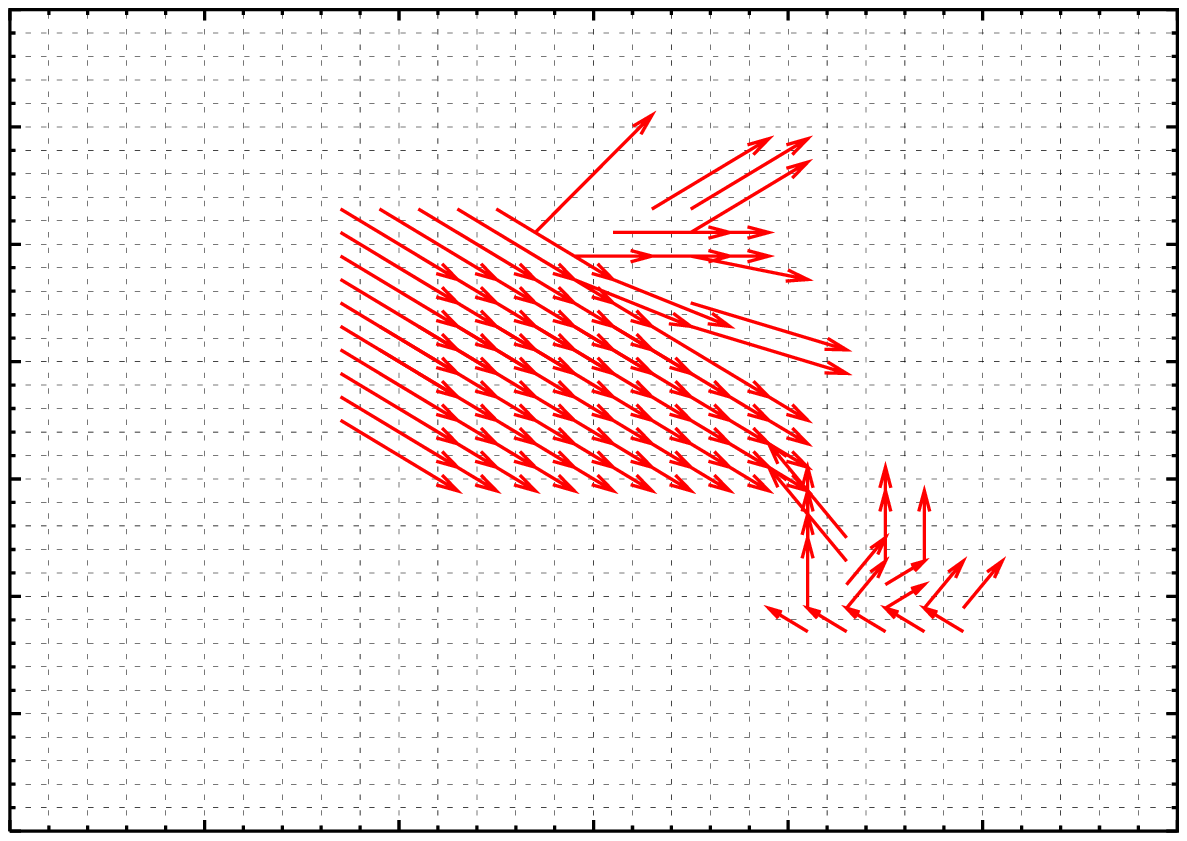} \hspace{-0.3cm}
\includegraphics[width=4.4cm]{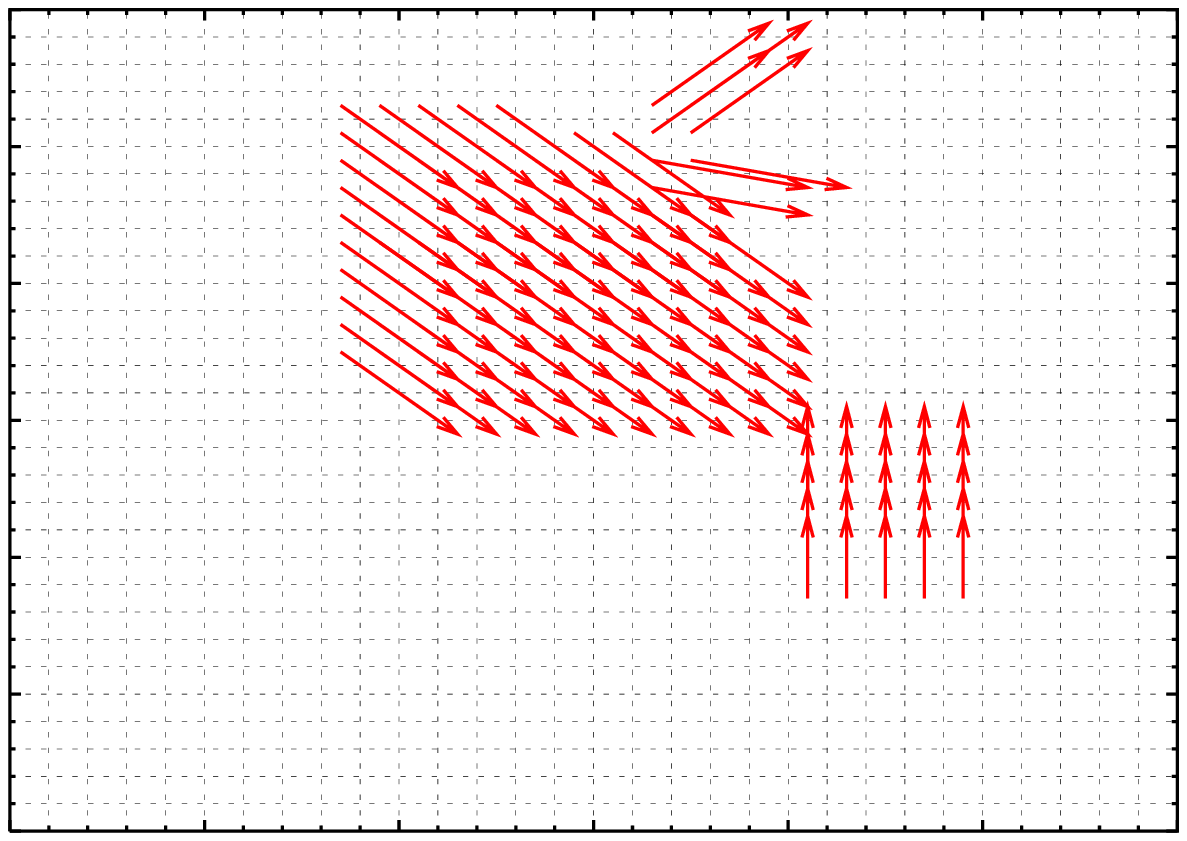}\\
\includegraphics[width=4.4cm]{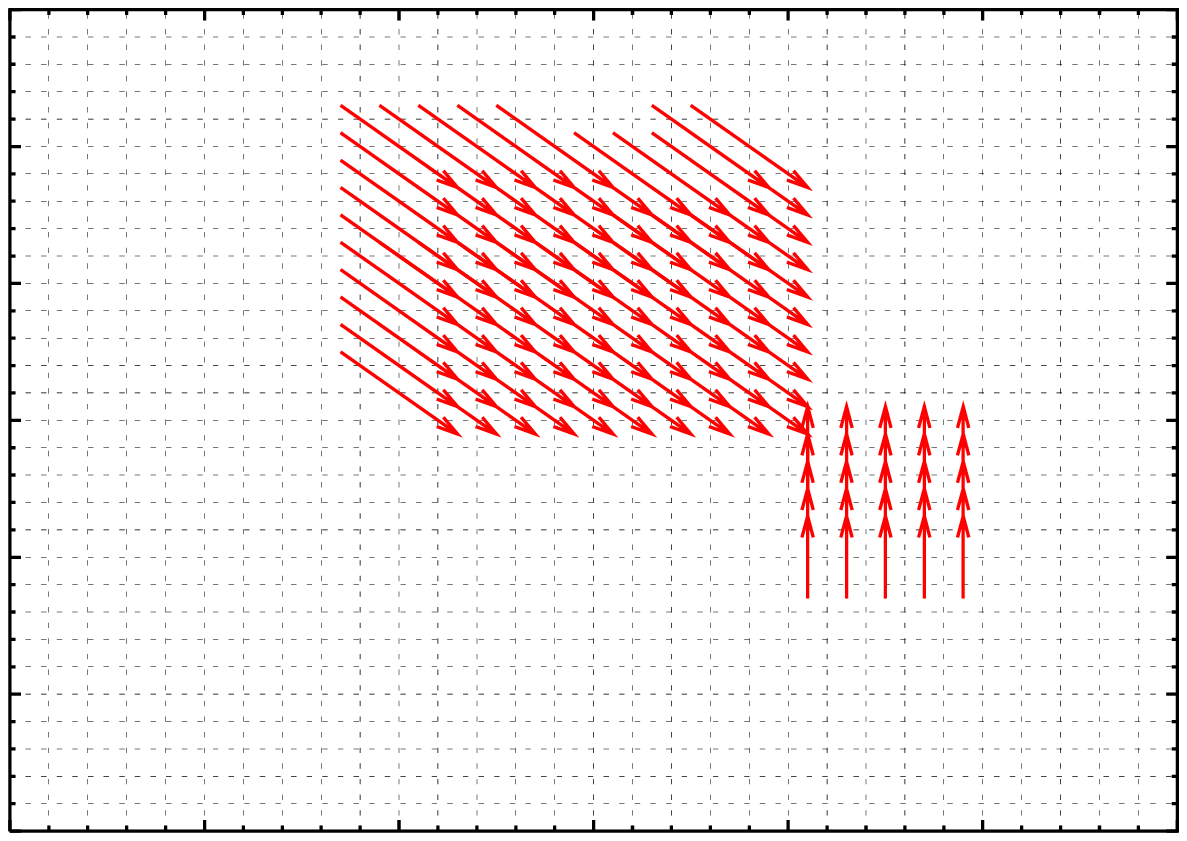} \hspace{-0.3cm}
\includegraphics[width=4.4cm]{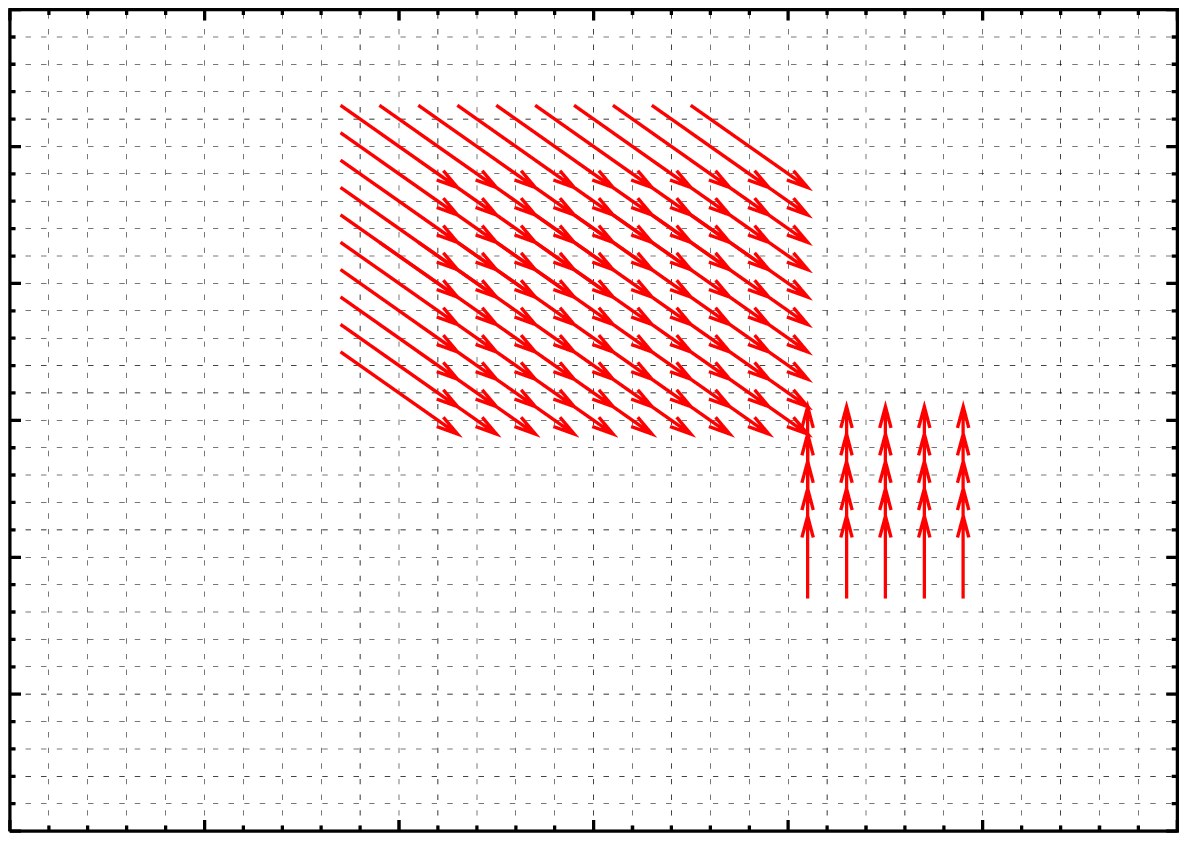}
\end{center}
\caption{
Typical snapshots of velocity fields 
obtained 
by the method of simple MCMC 
at time 
$t=0$ (upper left)，
$t=10$ (upper right), 
$t=20$ (lower left), 
$t=30$ (lower right)，
The case of 
$t=0$ corresponds to 
the result by 
Zhang and Hanouer (1995). }
\label{fig:jikken6-1_2}
\end{figure}

We show the results in Fig. \ref{fig:jikken6-1_2}. 
From these panels, we find that 
the resultant velocity fields at $t=30$ are much 
closer to the true fields than the result 
obtained by the hybridization. 
\mbox{}

\begin{figure}[ht]
\begin{center}
\includegraphics[width=5.5cm,angle=270]{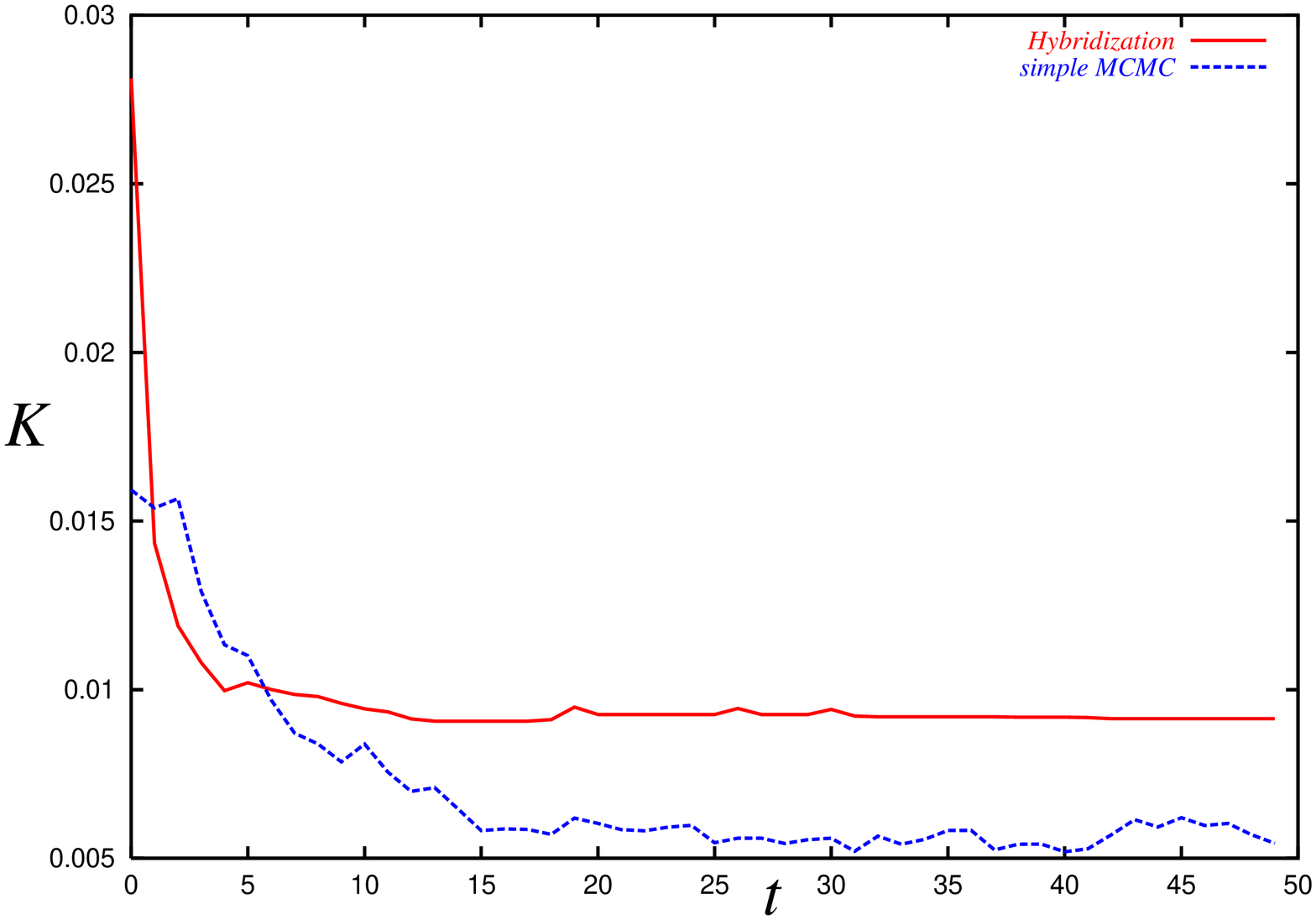} \\
\includegraphics[width=5.5cm,angle=270]{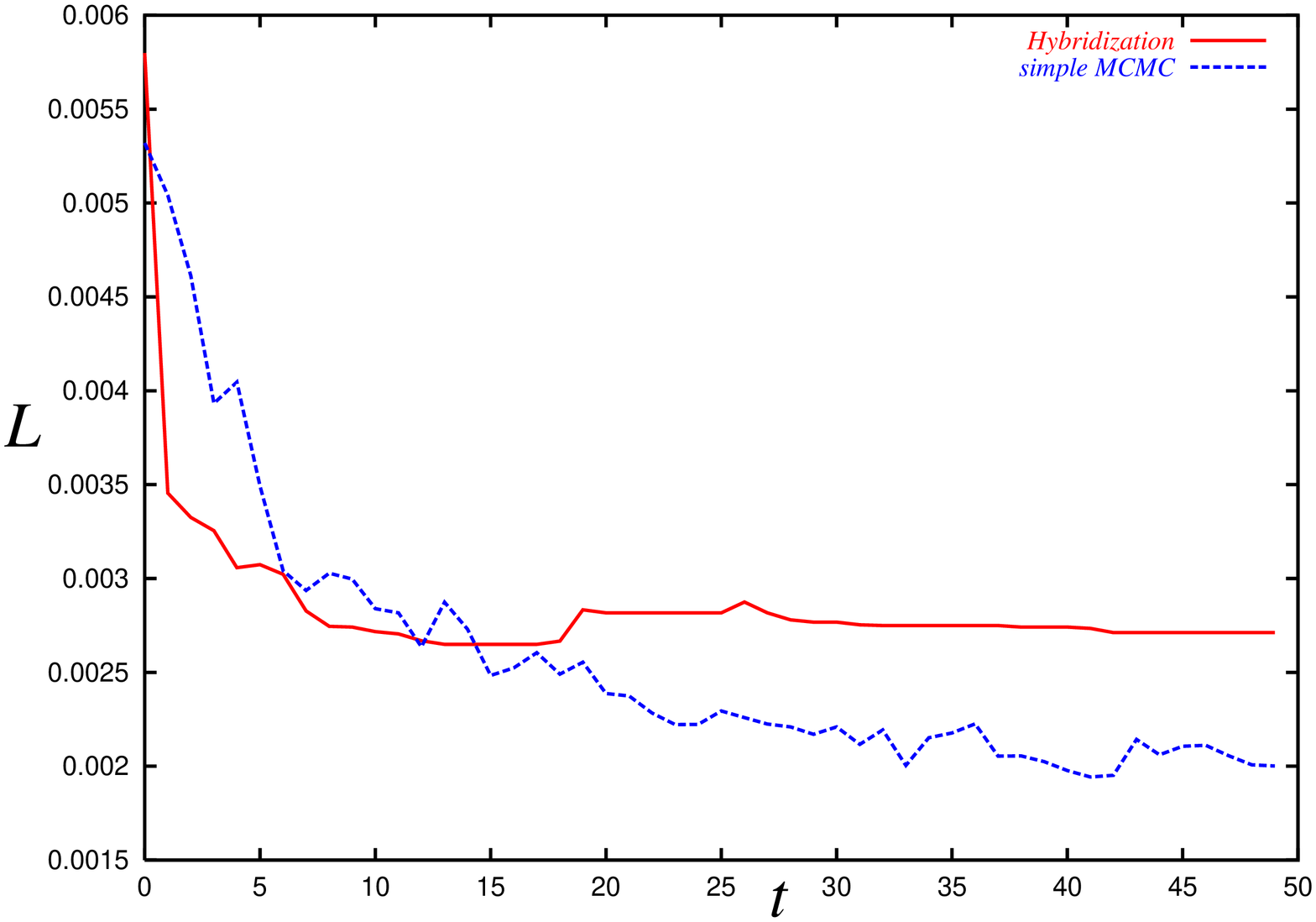} 
\end{center}
\caption{
The Euler step dependence of $K$ (upper panel) and $L$ (lower panel) for 
the hybridization (solid line) and simple MCMC (broken line). }
\label{fig:jikken6-1_6}
\end{figure}

We also evaluate 
the performance measures 
$K, L$ and 
compare the results with 
the results by the hybridization of mean-field approximation and 
MCMC in Fig. \ref{fig:jikken6-1_6}. 
From these two panels, we find that 
at the initial stage of the learning steps, 
the hybridization decreases 
the two kinds of errors 
very quickly, however, eventually 
the errors are saturated. 
On the other hand, 
the errors by  
the simple MCMC does not decreases so quickly at the initial stage, 
however, the resultant errors converge to lower values than those of 
the hybridization.  
\mbox{}

\begin{figure}[ht]
\begin{center}
\includegraphics[width=5.5cm,angle=270]{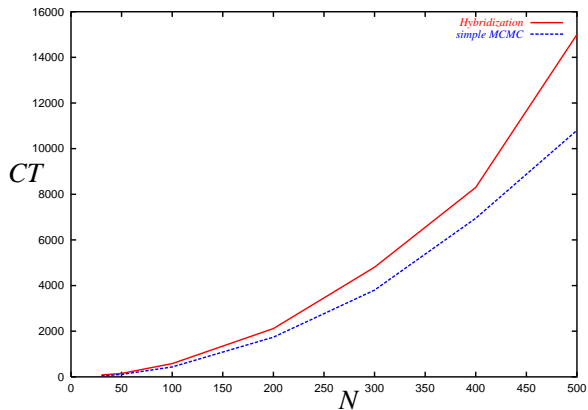} 
\end{center}
\caption{
Computational time (real CPU time) $CT$ [sec] for 
the hybridization (solid line) and simple MCMC (broken line) 
as a function of system size $N$.}
\label{fig:jikken6-1_7}
\end{figure}

We also compare the computational time until the convergence 
for hybridization and simple MCMC. 
The result is shown in Fig. \ref{fig:jikken6-1_7}. 
From this figure, we notice 
that the hybridization 
takes much longer time to proceed than the simple 
MCMC does because 
the Monte Carlo steps in the MCMC for each learning step 
$t$ are the same as the hybridization. 
\mbox{}

Finally we list the table to compare the hyper-parameters 
obtained by our methods and by Zhang and Hanouer (1995). 
\begin{table}[ht]
\begin{center}
\begin{tabular}{|c||c|c|c|}
\hline
\mbox{} & Zhang and Hanouer (1995) & 
Hybridization & simple MCMC \\
\hline
$\lambda_{s}$ & 2 & 2.3 & 2.5 \\
\hline
$B$ & 5 & 12.1 & 11.7 \\
\hline
$\lambda_{d}$ & 2.5 & 2.7 & 2.8 \\
\hline
$\beta_{d}$ & 4 & 3.8 & 3.7 \\
\hline
$\alpha_{l}$ & 200 & 232 & 220 \\
\hline
$T_{s}$ & 5 & 5 & 5 \\
\hline
\end{tabular}
\end{center}
\caption{\footnotesize 
Comparison of 
the resultant hyper-parameters.}
\label{tab:tb0}
\end{table}
We show the result in TABLE \ref{tab:tb0}. 
This table tells us that 
several parameters in Zhang and Hanouer (1995) 
are very close to ours or exactly the same as ours, however, 
some of the parameters are quite far from our results. 
This means that 
the ad-hoc choice by  Zhang and Hanouer (1995) 
is statistically (theoretically) incorrect and 
if one needs to choose statistically `proper'  
hyper-parameters `systematically', he (or she) should utilize the procedures provided by us in this paper.   
\section{Summary}
In this paper, we numerically examined a Bayesian mean-field approach 
with the assistance of the MCMC method 
to estimate motion velocity fields 
and probabilistic models simultaneously in consecutive digital images 
described by spatio-temporal Markov random fields. 
We found that our motion estimation is much better than the result obtained 
by Zhang and Hanouer (1995) in which the hyper-parameters are set to some ad-hoc values 
without any theoretical justification. 

Utilization of EM algorithm to determine the hyper-parameters 
by maximizing the marginal likelihood indirectly \cite{Inoue,TI},  
analytical evaluation of the average-case performance 
by making use of mathematically solvable MRFs such as 
Gaussian MRFs \cite{Tanaka2} or infinite range MRFs \cite{Inoue}, 
applying the Belief propagation \cite{Tanaka3} 
to compute the marginal probability 
in our framework are now on going and 
the results will be reported in the conference or elsewhere. 
\section*{Acknowledgment}
We were financially supported by Grant-in-Aid Scientific Research on
Priority Areas {\it `Deepening and Expansion of Statistical Mechanical Informatics
(DEX-SMI)'} of the MEXT No. 18079001. 
One of the authors (JI) was financially supported by 
INSA (Indian National Science Academy) -  JSPS 
(Japan Society of Promotion of Science)  Bilateral Exchange Programme. 
He also thanks Saha Institute of Nuclear Physics for their warm hospitality during 
his stay in India. 
We acknowledge Professor 
Tsuyoshi Horiguchi for drawing our attention to 
the reference \cite{Zhang} when 
he visited our research group in 2004.

\end{document}